\documentclass[10pt,journal,compsoc]{IEEEtran}
\usepackage{amsmath,amsfonts}
\usepackage{algorithmic}
\usepackage{algorithm}
\usepackage{array}
\usepackage{stfloats}
\usepackage{url}
\usepackage{verbatim}
\usepackage{graphicx}
\usepackage{cite}
\usepackage{nicematrix}
\usepackage{pifont}
\newcommand{\cmark}{\ding{51}}%
\newcommand{\xmark}{\ding{55}}%
\usepackage{hyperref}
\usepackage{booktabs}
\usepackage{bbm}
\usepackage{subfigure}
\usepackage[export]{adjustbox}
\usepackage{ragged2e}
\usepackage{xr}
\begin{document}

\title{Evidentially Calibrated Source-Free Time-Series Domain Adaptation with Temporal Imputation}

\author{Mohamed Ragab$^*$, Peiliang Gong$^*$, Emadeldeen Eldele, Wenyu Zhang, Min Wu, Chuan-Sheng Foo, Daoqiang Zhang, Xiaoli Li,~\IEEEmembership{Fellow,~IEEE,} and Zhenghua Chen$^\dag$

\thanks{$*$ These two authors contributed equally to this work. $\dag$ Corresponding author.}

\IEEEcompsocitemizethanks{
    \IEEEcompsocthanksitem Mohamed Ragab, Emadeldeen Eldele, Chuan-Sheng Foo, Xiaoli Li, and Zhenghua Chen are with the Institute for Infocomm Research (I$^2$R) and the Centre for Frontier AI Research (CFAR), Agency for Science, Technology and Research (A$*$STAR), Singapore (E-mail: mohamedr002@e.ntu.edu.sg, emad0002@ntu.edu.sg, foo\_chuan\_sheng@i2r.a-star.edu.sg, xlli@i2r.a-star.edu.sg, chen0832@e.ntu.edu.sg)
    \IEEEcompsocthanksitem Peiliang Gong is with the Key Laboratory of Brain-Machine Intelligence Technology, Ministry of Education, College of Computer Science and Technology, Nanjing University of Aeronautics and Astronautics, Nanjing 211106, China and the Insititute for Infocomm Research (I$^2$R), Agency for Science, Technology and Research (A$*$STAR), Singapore, 138632 (E-mail: plgong@nuaa.edu.cn)
    \IEEEcompsocthanksitem Wenyu Zhang, and Min Wu are with the Institute for Infocomm Research (I$^2$R), Agency for Science, Technology and Research (A$*$STAR), Singapore (E-mail: wenyu@i2r.a-star.edu.sg, wumin@i2r.a-star.edu.sg)
    \IEEEcompsocthanksitem Daoqiang Zhang is with the Key Laboratory of Brain-Machine Intelligence Technology, Ministry of Education, College of Computer Science and Technology, Nanjing University of Aeronautics and Astronautics, Nanjing 211106, China (E-mail: dqzhang@nuaa.edu.cn)
    \IEEEcompsocthanksitem This work is supported by the Agency of Science Technology and Research under its AME Programmatic (Grant No. A20H6b0151) and its Career Development Award (Grant No. C210112046).
}
}




\IEEEtitleabstractindextext{
\begin{abstract}
Source-free domain adaptation (SFDA) aims to adapt a model pre-trained on a labeled source domain to an unlabeled target domain without access to source data, preserving the source domain's privacy. While SFDA is prevalent in computer vision, it remains largely unexplored in time series analysis. Existing SFDA methods, designed for visual data, struggle to capture the inherent temporal dynamics of time series, hindering adaptation performance.
This paper proposes MAsk And imPUte (MAPU), a novel and effective approach for time series SFDA. MAPU addresses the critical challenge of temporal consistency by introducing a novel temporal imputation task. This task involves randomly masking time series signals and leveraging a dedicated temporal imputer to recover the original signal within the learned embedding space, bypassing the complexities of noisy raw data. During subsequent adaptation, the imputer network guides the target model to generate target features that exhibit temporal consistency with the source features. Notably, MAPU is the first method to explicitly address temporal consistency in the context of time series SFDA. Additionally, it offers seamless integration with existing SFDA methods, providing greater flexibility.
We further introduce E-MAPU, which incorporates evidential uncertainty estimation to address the overconfidence issue inherent in softmax predictions. To achieve that, we leverage evidential deep learning to obtain a better-calibrated pre-trained model and identify out-of-support target samples (those falling outside the source domain's support) by predicting them with higher entropy than source samples. During adaptation, the target classifier remains fixed while the feature extractor is trained to minimize the evidential entropy of out-of-support target samples. This is achieved by adapting the target encoder to map these samples to a new feature representation closer to the source domain's support. This fosters better alignment, ultimately enhancing adaptation performance.
Extensive experiments on five real-world time series datasets demonstrate that both MAPU and E-MAPU achieve significant performance gains compared to existing methods. These results highlight the effectiveness of our proposed approaches for tackling various time series domain adaptation problems.
\end{abstract}

}

\maketitle

\IEEEdisplaynontitleabstractindextext

\IEEEpeerreviewmaketitle

\section{Introduction}
\IEEEPARstart{D}{eep} learning has demonstrably achieved impressive results across various time series applications, including machine health monitoring, human activity recognition, and healthcare diagnostics. However, this success often hinges on the laborious and resource-intensive task of annotating large volumes of data. To alleviate this burden, unsupervised domain adaptation (UDA) has emerged as a promising technique. UDA leverages pre-labeled source data to train models on unlabeled target data, effectively bridging the distribution gap between the two domains \cite{da_survey}.

\begin{figure}[t]
\centering
\begin{tabular}{c}
\subfigure[]{\includegraphics[width=0.49\linewidth]{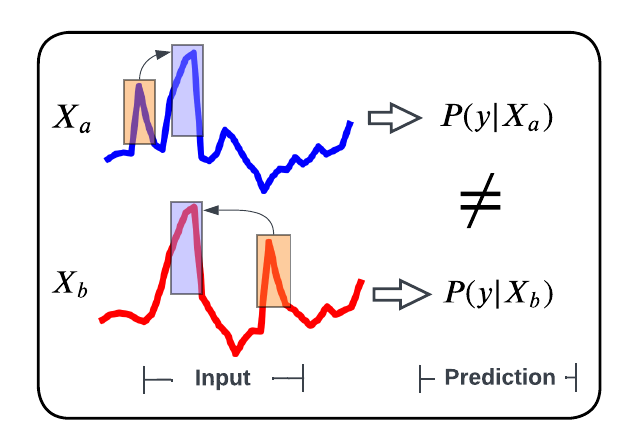} \label{fig_motivation_a}}
\subfigure[]{\includegraphics[width=0.49\linewidth]{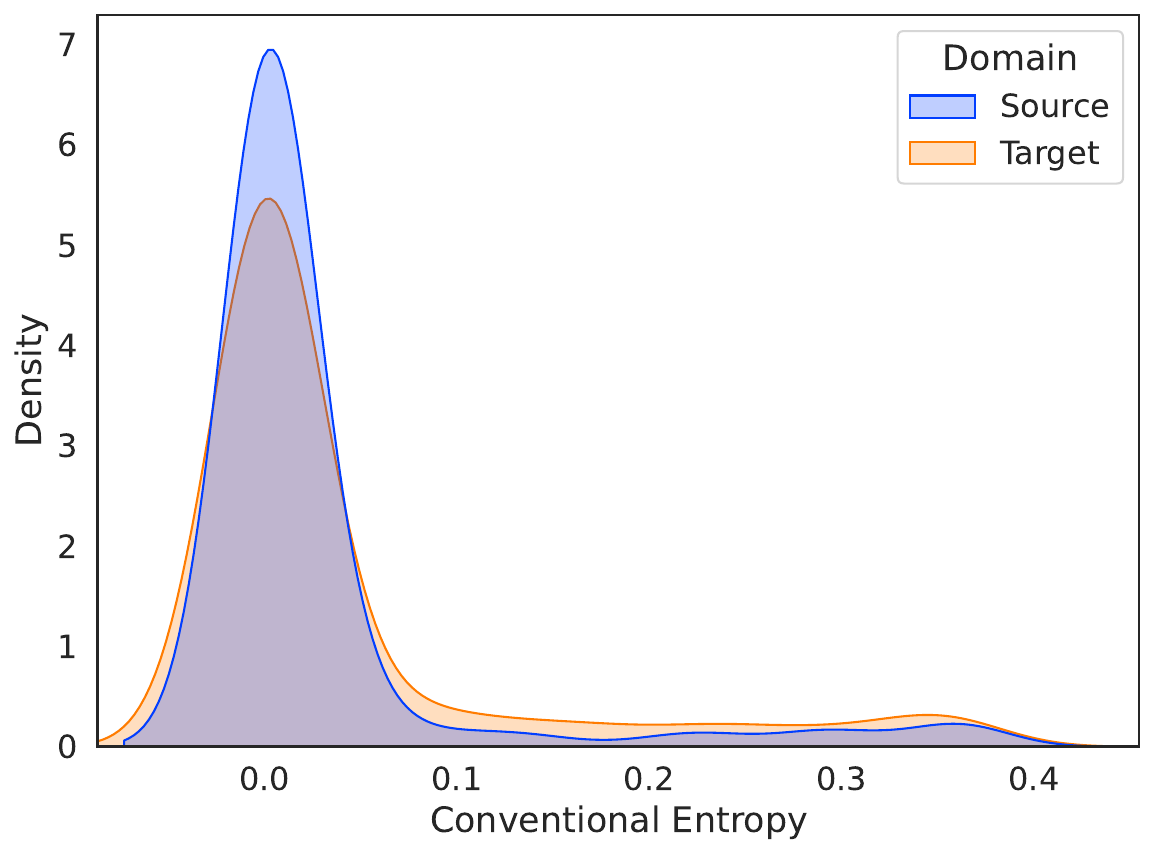} \label{fig_motivation_b}} \\
\subfigure[]{\includegraphics[width=0.49\linewidth]{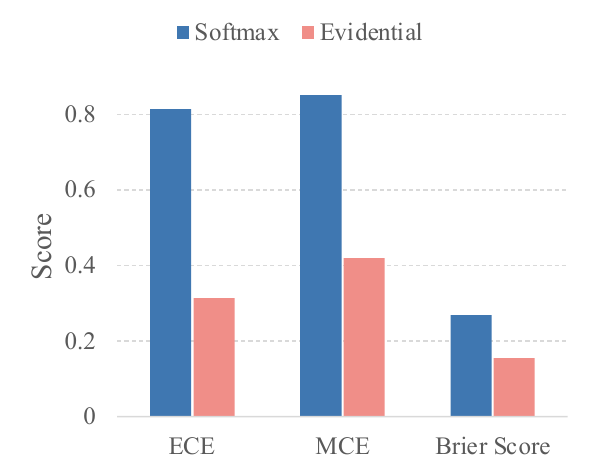} \label{fig_motivation_c}}
\subfigure[]{\includegraphics[width=0.49\linewidth]{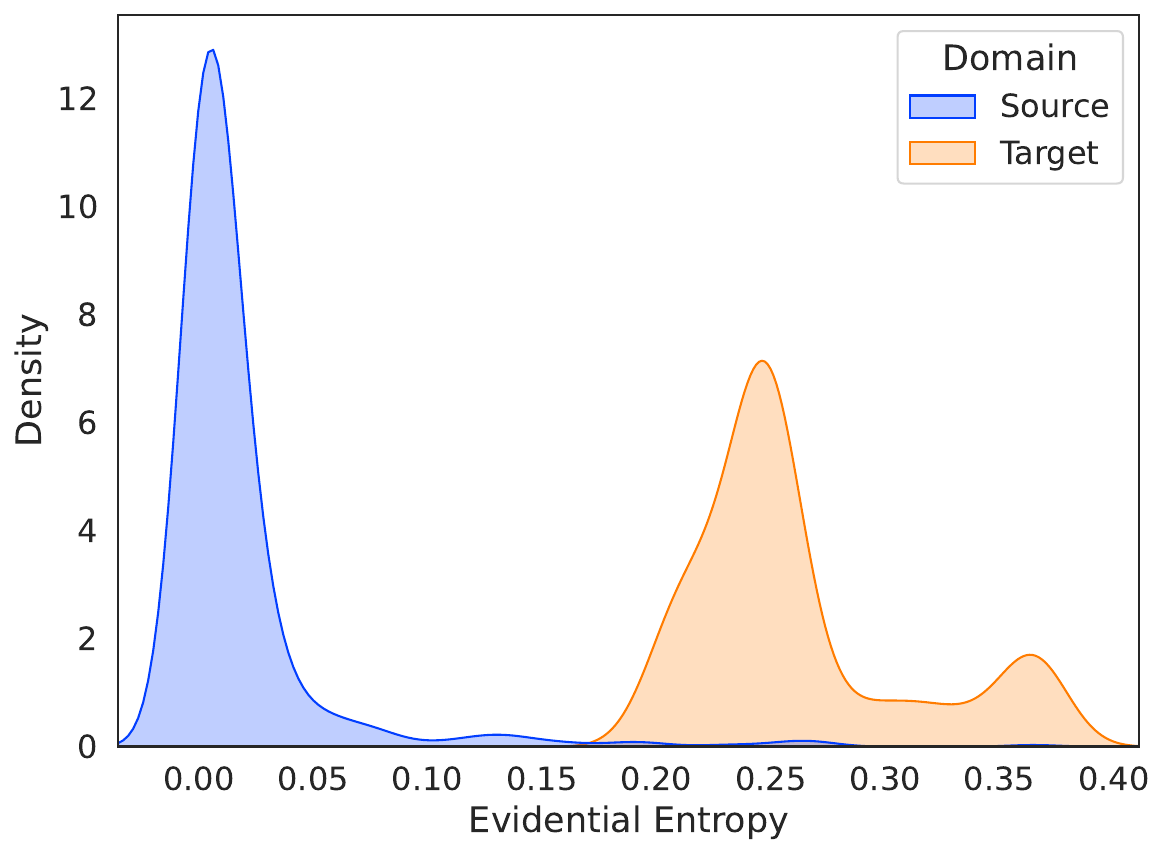} \label{fig_motivation_d}}

\end{tabular}
\caption{(a) How do the temporal relations matter in time series? Despite similarities in the values of the two signals, variations in the temporal position of their observations can result in different predictions. (b) Conventional entropy is not a reliable uncertainty estimator as it shows low uncertainty for both source and target domains. (c) The average calibration errors were calculated by using softmax scores and evidential learning on the UCIHAR dataset. Pre-training using evidential deep learning provides a better-calibrated model. (d) Our proposed approach with its evidential head can provide high uncertainty for the target domain and hence can detect the target samples outside the source support.}
\label{fig_motivation}
\end{figure}

Nevertheless, existing UDA methods have a significant limitation \textemdash they require access to the source data during the adaptation process. This requirement may not always be feasible due to data privacy regulations or other confidentiality concerns. To eliminate the dependency on source domain data during adaptation, source-free domain adaptation (SFDA) has been proposed \cite{sfda_survey}. In this setting, we can only access the weights of a model that has been pre-trained on the labeled source domain while adapting the unlabeled target domain. Recent years have witnessed significant development of SFDA methods for visual applications \cite{shot, sfda_kim,ss_sfda,sahoo_sfda,cpga}. A prevalent paradigm within these methods involves incorporating auxiliary tasks that leverage the inherent characteristics of visual data to enhance adaptation performance \cite{SHOT++,sl-sfda,ss_sfda}. However, applying these visual SFDA methods to time series may not effectively capture the temporal dynamics inherent within data.

One defining characteristic of time series data is their temporal dependencies, where values at different time steps are interrelated and significantly impact predictions \cite{temp_shift_1}. As demonstrated in Fig. \ref{fig_motivation_a}, the predictions of two signals with similar timestep-level observations but with a different temporal order can be completely distinct.
These temporal dynamics pose a significant challenge for UDA tasks, particularly when adapting between domains with temporal shifts. The challenge is further amplified in source-free settings, where the source data is unavailable during target domain adaptation. Consequently, the critical question is \textit{how to effectively adapt the temporal information in time series data with the absence of the source domain data}.

\begin{figure}[t]
\centering
\includegraphics[width=0.95 \linewidth]{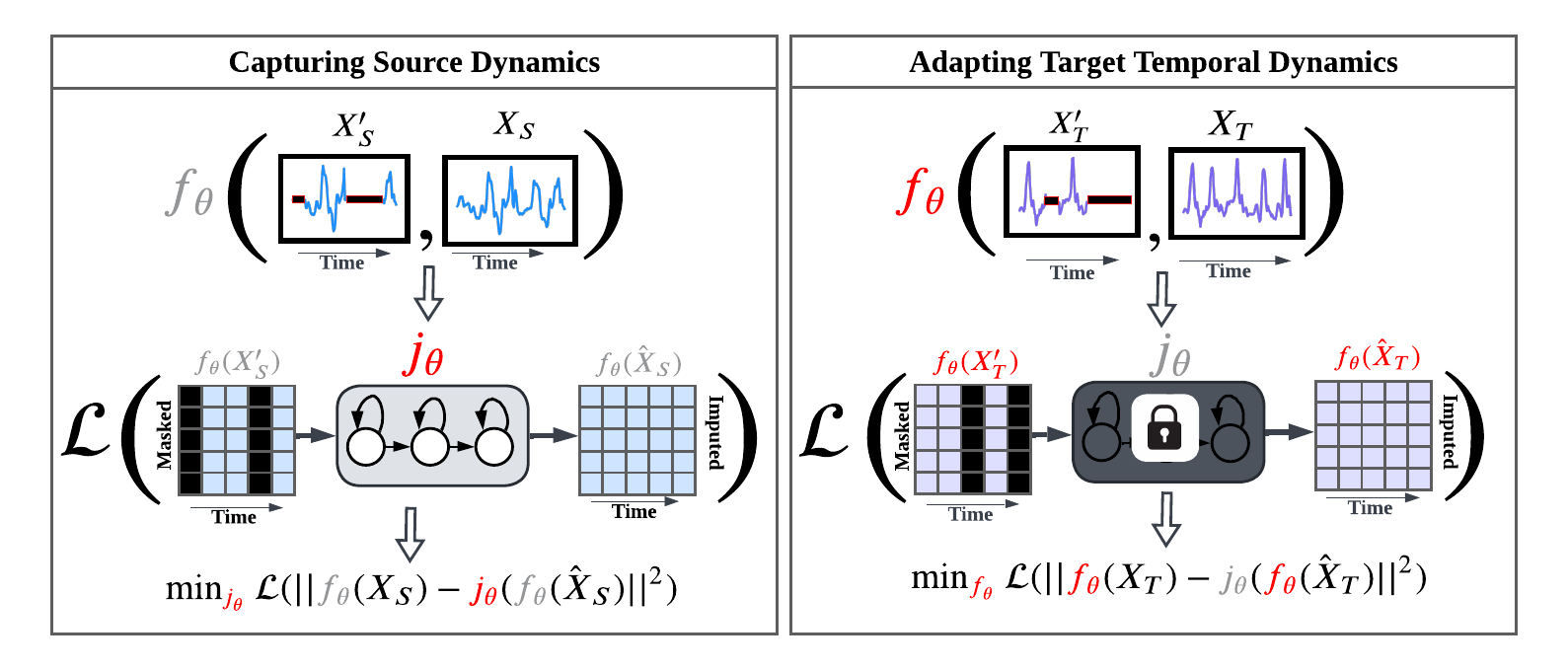}
\caption{Adaptation with Temporal Imputation. \underline{Left}: A temporal imputer network is trained to predict the full sequence from its masked version to capture the temporal information of the source domain. \underline{Right}: Once trained, the temporal imputer network guides the target model to produce temporally consistent features with the source domain. (Best in viewed in colors. Components in red color are trainable, while those in gray color are non-trainable).}
\label{fig_mapu_temporal_adapt}
\end{figure}

This work addresses the challenges associated with temporal adaptation in SFDA for time series data. We propose MAsk And imPUte (MAPU), a novel approach that leverages a temporal imputation task to effectively transfer knowledge about temporal dynamics from the source domain to the target domain. The core design of MAPU is illustrated in Fig.~\ref{fig_mapu_temporal_adapt}.
Specifically, our framework operates in two stages. 
In the first stage, the source model is trained using the standard cross-entropy loss. Further, we capture the temporal dynamics in time series data by applying a temporal masking operation to the input signals. Subsequently, both the masked and original signals are fed into an encoder network, generating the corresponding feature representations. Afterward, a dedicated temporal imputation network is trained within this feature space to recover the original signal from its masked counterpart, promoting smoother optimization during the imputation task.
This trained imputation network will be leveraged in the subsequent adaptation stage to guide the target model to generate features that can be accurately imputed by the source domain's imputation network, facilitating temporal consistency adaptation.

Despite the effectiveness of MAPU in tacking temporal adaptation for SFDA via its temporal imputation task, it inherits a limitation from current mainstream methods: a reliance on softmax probabilities and entropy for model regularization \cite{shot,sfda_kim,ss_sfda}. This approach assumes lower model confidence for out-of-distribution target samples. However, existing research demonstrates the vulnerability of softmax scores and conventional entropy to overconfidence issues, leading to high-confidence but incorrect predictions even on shifted data \cite{relu_confidence, energy_ood}. Fig.~\ref{fig_motivation_b} illustrates this limitation, showcasing how conventional entropy, calculated using softmax scores, fails to distinguish source and target domains due to overconfident predictions on target samples even in the presence of a clear domain shift. Consequently, minimizing entropy based on softmax predictions may not be ideal for adaptation, as many target samples are classified with low entropy.

To address this limitation, we propose Evidential-MAPU (E-MAPU), an extension of MAPU that incorporates evidential deep learning to accurately represent class prediction probabilities \cite{evident_dl}. Unlike the point estimates of the softmax, E-MAPU leverages a Dirichlet distribution \cite{dirichlet_dis} (the conjugate prior to the multinomial distribution) to model class probabilities, with the network's predictions determining the distribution's parameters. This approach fosters a better-calibrated pre-trained model and effectively tackles overconfidence by enabling clearer distinctions in confidence levels between in-distribution (source) and out-of-distribution (target) data. 
We posit that a better-calibrated model leads to improved adaptability. In Fig.~\ref{fig_motivation_c}, we compare the average calibration errors using softmax scores with those obtained using evidential learning. The results demonstrate significantly lower calibration errors for models with evidential learning, suggesting enhanced reliability and calibration.
Furthermore, Fig. \ref{fig_motivation_d} showcases that evidential entropy, derived from the estimated Dirichlet distribution, exhibits high confidence in the source domain and lower confidence in the target domain. To leverage this distinction, during adaptation, the target classifier remains fixed while the feature extractor is trained to minimize the evidential entropy of out-of-distribution target samples (those lying outside the source domain's support). This effectively compels the feature extractor to map these samples closer to the source domain's support, ultimately enhancing adaptation capabilities.

This journal paper extends our previous work on MAPU \cite{MAPU}. The key advancements are as follows: 
1) We extend the source domain pre-training stage of MAPU by incorporating evidential deep learning. This integration fosters a better-calibrated source model and addresses the inherent issue of overconfidence in softmax predictions.
2) A novel evidential loss is introduced to guide the target feature extractor. This loss steers out-of-support target samples towards a new representation that aligns more closely with the source domain's support, ultimately facilitating improved adaptation performance.
3) We add more datasets and conduct additional experiments to thoroughly evaluate the proposed MAPU and E-MAPU. These experiments demonstrate the effectiveness of the optimization strategy that leverages evidential uncertainty for enhanced model reliability and adaptation capabilities.
The principal contributions of this work are summarized as follows: 

\begin{itemize}
    \item To the best of our knowledge, we are the first to achieve the source-free domain adaptation for time series applications and provide two variants of the adaptation method. Both variants eliminate the requirement for source domain access during adaptation, offering a more practical solution in real-world scenarios.

    \item We introduce a novel temporal imputation task, specifically designed to ensure sequence consistency between source and target domains. This approach serves as a versatile foundation,  integrable with existing SFDA methods, effectively granting them temporal adaptation capabilities.
    
    \item We propose integrating evidential uncertainty as an objective function to mitigate distribution shift and enhance model robustness in SFDA. This strategy addresses the well-known issue of overconfidence inherent in standard softmax-based approaches.
    
    \item We conduct extensive experiments on the two proposed variants of our framework (MAPU and E-MAPU). The results convincingly demonstrate that both variants achieve significant improvements in adaptation performance on five real-world datasets, particularly for time series tasks.
\end{itemize}

\section{Related work}
\subsection{Time series Domain Adaptation}
Several approaches have been proposed to tackle the challenge of distribution shifts in time series data. These methods can be broadly classified into two categories: discrepancy-based methods and adversarial approaches.
Discrepancy-based approaches leverage statistical distances to align the feature representations of the source and target domains. For instance, AdvSKM employs the maximum mean discrepancy (MMD) distance with a hybrid spectral kernel to account for temporal relationships during adaptation \cite{dskn}. Another example is TSUDA, which learns the associative structure for time series data to align the source and target domains \cite{aaai_ts_uda}.
On the other hand, adversarial approaches exploit adversarial training to mitigate the distribution shift between the source and target domains. For instance, CoDATS exemplifies this strategy by employing adversarial training with multi-source data and weak supervision for human activity recognition \cite{codats}. DA\_ATTN combines adversarial training with an un-shared attention network to preserve domain-specific information during adaptation \cite{ts_da_attn}. Recently, the SLARDA method introduced an autoregressive adversarial training scheme to align the temporal dynamics across source and target domains \cite{slarda}.

While these existing methods achieve promising performance, they rely on the crucial assumption that source data is accessible during the adaptation step. In real-world scenarios, however, access to source data might be restricted due to privacy concerns or storage limitations. This necessitates the development of methods that can effectively adapt to new domains without requiring source data, a challenge addressed by our proposed approach.

\subsection{Source Free Domain Adaptation}   
To address the limitations of conventional UDA settings, a more practical scenario known as source-free domain adaptation (SFDA) has emerged. In SFDA, only a pre-trained source model is available during adaptation, as the source data itself is absent.
Several approaches have been proposed to achieve target domain adaptation without source data access \cite{SHOT++, sfda_de, sfda_kim,ss_sfda,sahoo_sfda, cpga, MSFUDA, trust_SFDA}. For example, without the labeled source data, one line of the research leverages the pre-trained source model to generate source-like data during adaptation \cite{cpga,3c_gan,sdda,sahoo_sfda}. Nevertheless, the data sampled from the target domain may not be fully representative of the underlying distribution in the source domain. Another line relies on adversarial training between multiple classifiers to achieve good generalization on target classes \cite{d_mcd,a2_net}. Additionally, a prevailing paradigm utilizes softmax scores or their corresponding entropy to prioritize confident samples for pseudo-labeling. This approach assumes the model should be highly confident on source samples and less confident on target samples \cite{shot,sfda_kim,ss_sfda}.

While these methods show promise, they have several limitations. Firstly, they are primarily designed for visual applications and may not effectively capture the temporal dynamics inherent in time series data. Our approach addresses this issue by introducing a novel temporal masking task, ensuring temporal consistency during domain adaptation. Secondly, existing methods often rely on softmax probabilities or entropy for adaptation, which can lead to overconfidence in classifying unseen target samples that have shifted distributions \cite{relu_confidence}. To overcome this issue, our approach leverages evidential deep learning to obtain more robust class probabilities, where the model's predictions exhibit high confidence for the source samples and low confidence for the out-of-distribution target samples. 

\subsection{Uncertainty Quantification for Deep Learning Models}
Uncertainty quantification for deep learning models is a rapidly growing research area. Various approaches have been proposed to estimate model uncertainty. For instance, Bayesian deep learning methods are designed to measure model trustworthiness \cite{kendall2017uncertainties,bao2020uncertainty}. However, these methods are often computationally expensive due to intractable posterior inference and the need for intensive sampling during uncertainty estimation. Alternative approaches, such as Monte Carlo (MC) dropout \cite{gal2016dropout} and deep ensembles \cite{lakshminarayanan2017simple}, have also been explored for uncertainty estimation. Nevertheless, these methods still rely on ensemble techniques or MC sampling, limiting their efficiency. 
Recently, evidential deep learning (EDL) \cite{evident_dl} has gained popularity for uncertainty quantification. EDL incorporates evidential theory into deep learning, interpreting categorical predictions as distributions through a Dirichlet prior placed on class probabilities. Compared to existing methods for uncertainty estimation, EDL only requires a single forward pass, significantly reducing computational costs. Due to this benefit, EDL has been leveraged and shown promise for various tasks, including out-of-distribution (OOD) detection in visual tasks \cite{chen2022evidential_TNT,evident_dl,zhao2019quantifying}, OOD node detection in graph neural networks \cite{zhao2020uncertainty_GKDE}, and trustworthy long-tailed classification \cite{li2022trustworthy_TCL}.

In contrast to existing works, our approach is the first to utilize evidential uncertainty for source-free domain adaptation on time series data. We hypothesize that calibrated class probabilities, obtained through EDL, enable more accurate identification of in-distribution source samples and out-of-distribution target samples. To achieve this, we employ EDL to obtain robust class probabilities and leverage evidential entropy to guide domain alignment. This represents a novel contribution to the field of domain adaptation.

\section{Methodology}
We aim to address the challenge of the source-free domain adaptation problem for time series data, where access to the source data is strictly prohibited during the adaptation phase. In particular, we consider the scenario aligned with the vendor-client paradigm \cite{vc1,vc2,vc3,vc4}, which allows the influence of the source pretraining stage. This setting reflects real-world applications where collaboration exists between entities, but data sharing is restricted by privacy, security, or regulations. Given a labeled source domain $\mathcal{D}_S = \{{X}_S^i, {y}_S^i\}_{i=1}^{n_S} $, where ${X}_S \in \mathcal{X}_S$ can be a uni-variate or multi-variate time series data with a sequence length $L$, while $y_S \in \mathcal{Y}_S$ represents the corresponding labels. In addition, we have an unlabeled target domain $\mathcal{D}_T= \{{X}_T^i\}_{i=1}^{n_T} $, where ${X}_T \in \mathcal{X}_T$, and it also shares the same label space with $\mathcal{D}_S$. We follow the common UDA assumption of differing marginal distributions across domains (i.e., $P(X_S) \neq P(X_T)$) while maintaining conditional distribution similarity (i.e., $P(y_S|X_S) \approx P(y_T|X_T)$).

\begin{figure}[t]
\centering
\includegraphics[width=0.95 \linewidth]{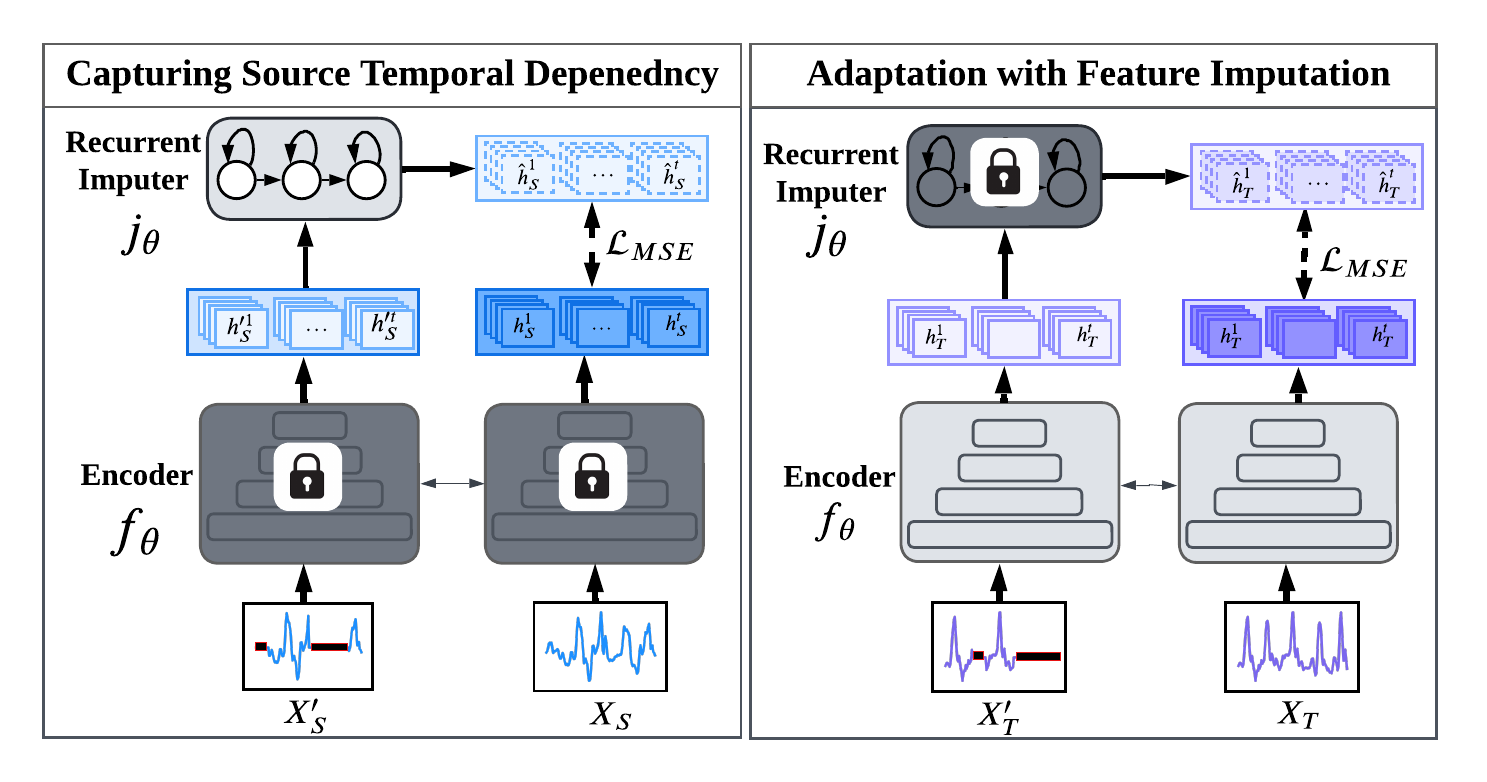}
\caption{Adaptation with Temporal Imputation for time series data. \underline{Left}: The pretraining stage of the temporal imputer network $j_\theta$ to capture the temporal dynamics of the source domain. First, we perform random masking across the time dimension of the source signal. Given the original source signal $X_S$ and its temporally masked signal $X_S^\prime$, the encoder network $f_\theta$ is used to generate the corresponding latent features $H_S$ and $H_S^\prime$ respectively. Subsequently, $j_\theta$ is updated to produce imputed features $\hat{H}_S$ from masked features $H_S^\prime$ using the mean square error loss. \underline{Right}: The adaptation stage of the encoder network on the target domain data. The encoder $f_\theta$ is updated to produce source-like features that are imputable by the pre-trained $j_\theta$.}
\label{fig_key_idea_mapu}
\end{figure}

\begin{figure*}[t]
\centering
\includegraphics[width=1.0 \linewidth]{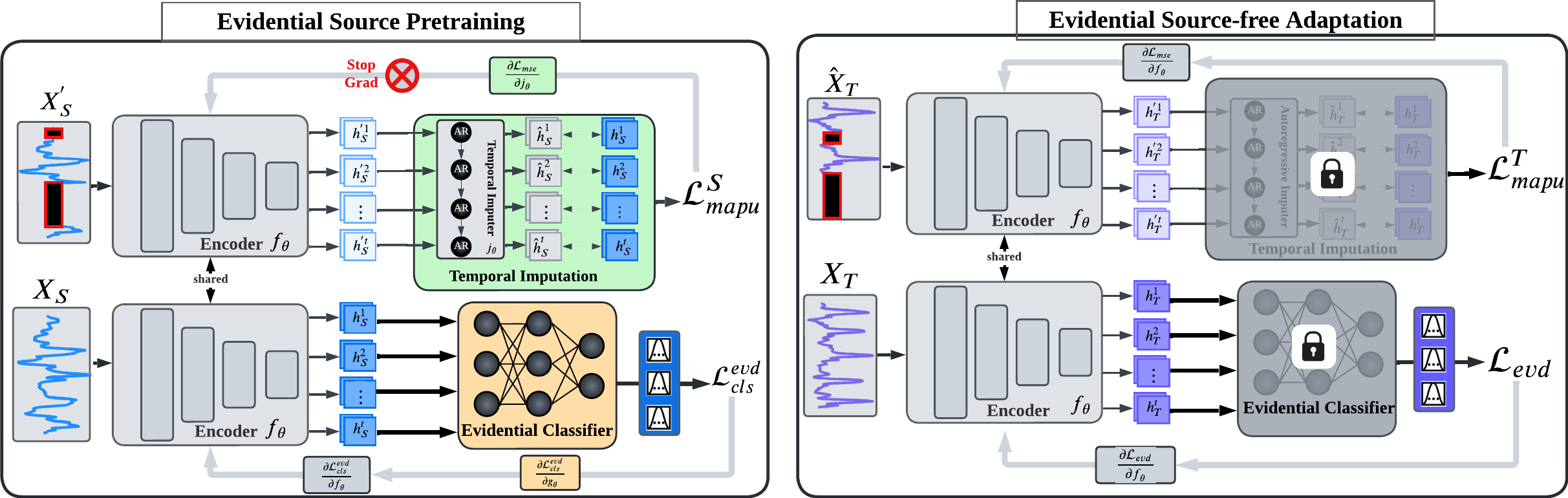}
\caption{The illustration of Evidential Source Free Adaptation with Temporal Masking (Best viewed in colors). \underline{Left}: In the pretraining stage, the source model is trained using evidential cross-entropy loss $\mathcal{L}_{cls}^{evd}$, and the temporal imputer network is trained using $\mathcal{L}_{mapu}^S$ to impute the features of the masked signals and capture the source temporal information on the feature space. \underline{Right}: In the adaptation stage, the target model is jointly trained with both an evidential adaptation loss $\mathcal{L}_{evidence}$ and our temporal imputation loss $\mathcal{L}_{mapu}^T$ to perform the adaptation while ensuring temporal consistency with the source features. The $h_S^\prime$ and $h_T^\prime$ are the feature representations of the input signals extracted by the encoder $f_\theta$ in the source and target domains, respectively.}
\label{fig_main_mapu_plus_plus}
\end{figure*}

We present a novel framework, MAPU, to achieve source-free adaptation on time series data while considering the temporal dependencies across domains. The core design of MAPU lies in the temporal imputation task, as illustrated in Fig. \ref{fig_key_idea_mapu}. Given the input signal and a temporally masking signal, the framework operates in two stages. First, an autoregressive network, i.e., the ``imputer network'' is trained on the source domain using a temporal imputation task. This task involves reconstructing the original signal from a temporally masked version, effectively capturing the temporal dynamics within the source domain. Then, during adaptation, the pre-trained imputer network guides the target encoder to generate temporally consistent target features. 
To further enhance model reliability and adaptation capabilities, we propose E-MAPU addressing the issue of overconfidence in softmax predictions by incorporating evidential uncertainty learning into the MAPU framework, as illustrated in Fig. \ref{fig_main_mapu_plus_plus}. Specifically, during source pre-training, we employ an evidential learning framework based on Bayesian risks with evidential cross-entropy loss. This replaces the conventional cross-entropy loss function, resulting in better-calibrated class probabilities. In the adaptation phase, we optimize the target model to minimize the evidential entropy (instead of conventional entropy) on the target samples that lie outside the source support, adapting the out-of-distribution target samples towards the source domain. Next, we elaborate on each component in more detail.

\subsection{Preliminaries}
Most of the existing deep learning methods rely on softmax to generate class probabilities. However, softmax offers only point estimates of the output distribution and suffers from overconfidence, particularly for unseen samples \cite{ovadia2019can, relu_confidence}. To address this, we incorporate evidential learning, which leverages a Dirichlet distribution imposed over the model predictions. This approach infers a calibrated predictive uncertainty, effectively alleviating the overconfidence issue inherent in softmax probabilities \cite{evident_dl}. Quantifying uncertainty within a Dirichlet distribution for multi-class classification problems is facilitated by subjective logic theory. This framework allows us to associate Dirichlet parameters with both mass and belief \cite{subjective_logic}. Specifically, for a $K$-class classification problem, a belief mass $\{b_k\}_{k=1}^{K}$ is assigned to each class, and an uncertainty mass $u$ is assigned to the entire set of classes. Both belief and uncertainty masses are non-negative and sum to 1:
\begin{align}
u +  \sum_{k=1}^{K} b_k = 1.  
\end{align}

For a given sample $x_i$, the evidential learning framework calculates the uncertainty mass and belief mass associated with the $k^{th}$ class. These are defined as,
\begin{align}\label{dir}
    u^i = \frac{K}{S^i} \text{ and } b_k^i = \frac{e_k^i}{S^i},
\end{align}
where $\textbf{e}_i = ln(1+e^{\textbf{h}_i})$ represents the collected evidence for the $K$-class problem according to Dempster-Shafer Theory (DST) \cite{dempster_shafer_theory}. Here, $\textbf{h}_i \in \mathbb{R}^{K}$ denotes the model's predicted scores for the sample. $S^i=\sum_{k=1}^{K} \alpha_k^i$ is the Dirichlet strength, calculated by summing the concentration parameters for all classes. The estimated evidence is then used to determine the concentration parameters of the Dirichlet distribution Dir$(P|\alpha)$. Specifically, $\alpha_k^i = e_k^i + 1$ represents the concentration parameter for the $k^{th}$ class.

As shown in Eq. \ref{dir}, the lower the evidence collected for the $k^{th}$ class the higher the uncertainty mass imposed and vice versa. By considering the belief mass as subjective opinion, we estimate the class probabilities as the mean of the concentration parameters $\alpha_k^i$ for the Dirichlet distribution,
\begin{align}
    \hat{p}_k^i = \frac{\alpha_k^i}{\sum_{k=1}^{K} \alpha_k^i}.
\end{align}

\subsection{Source Model Generation}
\subsubsection{Pre-training with conventional cross-entropy}
We consider learning a deep source classification model, which consists of a feature encoder network $f_\theta: \mathcal{X}_S \rightarrow \mathcal{H}_S$ and a classifier network $g_\theta:\mathcal{H}_S \rightarrow \mathcal{Y}_S$. We achieve this by minimizing the cross-entropy loss function with the label smoothing technique \cite{label_smoothing}. Label smoothing introduces a degree of softness into the one-hot encoded labels, encouraging the model to learn more robust representations. The loss function is formulated as follows,
\begin{align}
    \min_{f_{\theta}, g_{\theta}}\mathcal{L}_{cls} 
    &= - \Bbb{E}_{(x_s, y_s) \sim (\mathcal{X}_s, \mathcal{Y}_s)} \sum_{k=1}^{K} y_{k}^{ls} \log {\delta_{k}(g_{\theta}(f_{\theta}(x_s)))},
\end{align}
where $\delta_{k}(\cdot)$ denotes the softmax function, $y_{k}^{ls}$ represents the $k^{th}$ element of the smooth label $y_{k}^{ls} = (1-\eta)y_k + \eta / K$, $\eta$ is the smoothing coefficient which is empirically set to $0.1$,  and $y_k$ is the one-hot encoding of label.

\subsubsection{Capturing source temporal dynamics with temporal imputation}
Effective adaptation to target time series domains necessitates considering the inherent temporal relationships within the source domain. Relying solely on cross-entropy loss for source network training may overlook this crucial aspect. To address this, we propose a novel temporal imputation task. This task challenges the model to learn temporal dependencies by reconstructing the original input signal $X$ from a temporally masked version $X^\prime$ within the feature space. First, the temporal masking process involves dividing the input signal into blocks along the time dimension. Subsequently, a random selection of these blocks is set to zero, generating the masked signal $X^\prime$. This masking is applied to both source and target domains. By learning to impute the missing information from surrounding blocks, the model is encouraged to capture the temporal dynamics present in the signal. 

Then, the imputation task is performed by an imputer network $j_\theta$ that takes the masked signal and maps it to the original signal. The input signal $X_S$ and masked signal $X_S^\prime$ are first transformed into their corresponding feature representations $H_S$ and $H_S^\prime$ by the encoder $f_\theta$. The imputer network aims to learn a mapping such that $\hat{H}_S^\prime = j_\theta(f_\theta(X_S^{\prime})) \rightarrow H_S = f_\theta(X_S)$, where $\hat{H}_S^\prime$ denotes the imputed signal. To achieve this, the network is trained by minimizing the mean squared error between the features of the original and imputed signals, which can be formulated as:
\begin{align}
    \min_{j_{\theta}}\mathcal{L}_{mapu}^S = \frac{1}{n}\sum_{i=1}^{n_S}\left\|f_\theta(X_S^i) - j_\theta(f_\theta(X_S^{\prime i}))\right\|_2^2,
\end{align}
where $H_S = f_\theta(X_S^i)$ represents the latent features of the original signal, $\hat{H}_S = j_\theta(f_\theta(X_S^{\prime i}))$ is the output of the imputer network, and $n_S$ is the total number of source samples. 

Eventually, the pre-training of the source model for the proposed MAPU is completed by jointly optimizing the cross-entropy loss and the temporal imputation loss, which is formulated as,
\begin{align}
    \mathcal{L}_{src} = \mathcal{L}_{cls} + \mathcal{L}_{mapu}.
\end{align}

Notably, the loss associated with the temporal imputation task does not propagate to the encoder model $f_{\theta}$. As a result, the encoder can be trained exclusively with the conventional cross-entropy loss, ensuring that the imputation task does not negatively impact the pretraining performance.

\subsubsection{Pre-training with evidential cross-entropy} Furthermore, in E-MAPU, to provide a better-calibrated pre-trained model with uncertainty quantification capability, we train an evidential network $u_\theta:\mathcal{H}_s \rightarrow \mathcal{U}_s$ on the source features $\mathbf{h}_s$ to learn the task and provide a robust uncertainty measure. Specifically, we leverage the evidential deep learning work based on Bayes risk with a cross-entropy loss instead of the conventional cross-entropy loss in MAPU to train the evidential head \cite{evident_dl}:
\begin{align}
    \mathcal{L}_{cls}^{evd}
    &= \mathcal{L}_{CE}^{evd} + \lambda_{t} \mathcal{L}_{KL}.
\end{align}

Here, the $\mathcal{L}_{CE}^{evd}$ denotes evidential cross-entropy that ensured that the model assigned higher confidence scores to correct predictions compared to other classes. Conversely, the $\mathcal{L}_{KL}$ denotes Kullback-Leibler (KL) divergence term that penalized the model for assigning high confidence to incorrect predictions. A balance factor $\lambda_{t}=min(\frac{t}{10}, 1) \in [0, 1]$ was dynamically increased during training to prevent the model from focusing excessively on minimizing the KL divergence in early stages, $t$ is the index of the current training epoch. This early focus could hinder the exploration of the parameter space and potentially lead the network to converge on a flat, uninformative distribution. Specifically, 
\begin{align}
    \mathcal{L}_{CE}^{evd} 
    &= \int \left[ \sum_{k=1}^K -{y_k} \log {p_k} \right] \frac{1}{B(\alpha)} \prod_{k=1}^{K} p_{k}^{\alpha_{k}-1} dp  \\
    &= \sum_{k=1}^K y_k \bigg( \psi (S) - \psi ({\alpha}_k) \bigg),
\end{align}
where $\psi(\cdot)$ is the digamma function, $S=\sum_{k=1}^{K} \alpha_{k}$ is the Dirichlet strength, $\alpha_k = e_k + 1$ is parameter of the estimated Dirichlet distribution $D(P|\alpha)$, and $e_k = \phi(u_\theta(f_\theta(x)))$, where $\phi(\cdot)$ is the softplus function. Besides, the KL divergence regularity term is as shown follows,
\begin{align}
    \mathcal{L}_{KL} 
    &= KL \left[ D(p|\hat{\alpha}) \parallel D(p|<1,....,1>)\right] \\
    &= \log \bigg( \frac{\Gamma (\sum_{k=1}^{K} \hat{\alpha}_{k})}{\Gamma(K) \prod_{k=1}^{K} \Gamma(\hat{\alpha}_{k})} \bigg) \\
    &+ \sum_{k=1}^{K}(\hat{\alpha}_{k} - 1)\left[ \psi(\hat{\alpha}_{k} - \psi(\hat{S}))\right],
\end{align}
where $\Gamma(\cdot)$ is the gamma function, $\psi(\cdot)$ is the digamma function, $\hat{\alpha}_{k}=y_k+(1-y_k)\alpha_{k}$ is adjusted Dirichlet parameters after remove the correct evidence, and $\hat{S}=\sum_{k=1}^{K} \hat{\alpha}_{k}$.

In this manner, the pre-training of the source model for E-MAPU can be accomplished by jointly optimizing the evidential loss and the temporal imputation loss as follows,
\begin{align}
    \mathcal{L}_{src} = \mathcal{L}_{cls}^{evd} + \mathcal{L}_{mapu}.
\end{align}

\subsection{Target Domain Adaptation}
\subsubsection{Temporal adaptation with feature imputation}
During adaptation, the goal is to train the target encoder network $f_\theta$ to produce target features temporally consistent with the source features. This is achieved by leveraging the pre-trained imputer network, $j_\theta$. The encoder network extracts latent feature representations from both the original target sample $X_T$ and its masked version $X_T^\prime$. Subsequently, the fixed imputer network attempts to reconstruct the original target features from the masked features. However, due to inherent domain differences, the source imputer might not perfectly reconstruct the target features. Thus, we update the target encoder to produce target features that can be accurately reconstructed by the imputer network. This optimization process can be formulated as,
\begin{align}
\label{temporal_adapt_eq}
\min_{f_\theta}\mathcal{L}_{mapu}^T = \frac{1}{n}\sum_{i=1}^{n_T}\left\|f_\theta(X_T)^i - j_\theta(f_\theta(X_T^\prime)^i))\right\|_2^2,
\end{align}
where $H_T= f_\theta(X_T)$ are the original target features, $\hat{H}_T = j_\theta(f_\theta(X_T^\prime))$ are the adapted target features produced by the imputer network to minimize the mean square error loss, and $n_T$ is the total number of target samples. 

Notably, only the target encoder network is optimized during adaptation. This allows it to generate target features that can be effectively imputed by the fixed source-pretrained imputer network. By reducing the imputation loss, the adapted target features are encouraged to maintain temporal consistency with the source features. Algorithm \ref{alg:temp_imput} outlines the adaptation procedure via temporal imputation. The process begins by first constructing a temporally masked version of the input target sample, denoted as $X_T^{\prime}$. Subsequently, the pre-trained source encoder is employed to extract latent features from both the original target signal and its masked version. These features are represented as $H_T$ and $H_T^{\prime}$, respectively. Finally, the target encoder network is updated through backpropagation to ensure the source pre-trained imputer network can effectively recover the masked signal's features. This optimization process minimizes the mean squared error loss, as defined in Eq. \ref{temporal_adapt_eq}.

\begin{algorithm}[tb]
\caption{Adaptation with Temporal Imputation}
   \label{alg:temp_imput}
\begin{algorithmic}[1]
\STATE \textbf{Input:} Target sample $X_T$, source pretrained encoder $f_\theta$, source classifier $g_\theta$, imputer network $j_\theta$ 

\STATE \textbf{Initialize:} Construct temporally masked version of $X_T$: $X_T^{\prime}$

\STATE \textbf{Extract:} Latent feature representations: $H_T = f_\theta(X_T)$, $H_T^{\prime} = f_\theta(X_T^{\prime})$

\STATE \textbf{Impute:} Masked features using imputer network: $\hat{H}_T = j_\theta(H_T^{\prime})$

\STATE \textbf{Update:} Encoder $f_\theta$ to produce target features that can be accurately reconstructed by $j_\theta$ using Eq. \ref{temporal_adapt_eq}

\STATE \textbf{Output:} Updated encoder $f_\theta$ 
\end{algorithmic}
\end{algorithm}

\subsubsection{Domain adaptation with evidential uncertainty}
We propose a target adaptation strategy that incorporates evidential uncertainty estimation by leveraging the pre-trained source encoder $f_\theta$ and the evidential network $u_\theta$. This process aims to map all target samples closer to the source domain's support, such that the frozen evidential network assigns low uncertainty estimates to them. For each target sample $x_t^i$, we first extract the corresponding features $h_t^i = f_\theta(x_t^i)$ using the source encoder. Subsequently, the evidential network $u_\theta$ is applied to these features to estimate the uncertainty associated with the target sample. Ideally, the uncertainty network should assign high uncertainty values to target samples that fall outside the source domain's support, effectively discerning them from source samples. After the detection of high-uncertainty target samples, the weights of the evidential network $u_\theta$ are frozen. We then optimize the encoder network $f_\theta$ to learn new target features that exhibit low uncertainty. To achieve this, we minimize the following evidential loss function during target adaptation,
\begin{align}
    \min_{f_\theta}\mathcal{L}_{evd} = \gamma_1 \mathcal{L}_{entropy}^{evd} + \gamma_2 \mathcal{L}_{diversity}^{evd} + \gamma_3 \mathcal{L}_{sl}^{evd}, 
\end{align}
where $\gamma_1$, $\gamma_2$, and $\gamma_3$ are hyperparameters used to weight the individual loss components within the overall evidential loss. 

Here, the $\mathcal{L}_{entropy}^{evd}$ denotes evidential entropy loss. Minimizing the evidential entropy loss encourages the target encoder to map out-of-support samples to new feature representations that lie within the source domain's support. This is achieved by,
\begin{align}
    \mathcal{L}_{entropy}^{evd} = \sum_{i=1}^{n_t} \sum_{k=1}^{K} \hat{p}_{k}^{i} \log (\hat{p}_{k}^{i}),
\end{align}
where $p_k^i = \frac{\phi(u_\theta(f_\theta(x_t^i)))+1}{S}$ is the predicted Dirichlet probability for the $k^{th}$ class of sample $i$. 
Concurrently, maximizing the diversity loss $\mathcal{L}_{diversity}^{evd}$ promotes distinctiveness among the evidential probabilities. This is achieved by maximizing the distance between the mean value and the probabilities of all individual samples \cite{shot}: 
\begin{align}
    \mathcal{L}_{diversity}^{evd} 
    &= - \sum_{k=1}^{K}\sum_{i=1}^{n_t} \hat{p}_k^i \log (\hat{p}_k^i),
\end{align}
Furthermore, the self-supervised evidential loss $\mathcal{L}_{sl}^{evd}$ incorporates evidential learning based on Bayes risk with the cross-entropy loss on the target domain, formulated as,
\begin{align}
    \mathcal{L}_{sl}^{evd}
    &= \mathcal{L}_{sl\_CE}^{evd} + \lambda_t\mathcal{L}_{KL} \\
    &= \sum_{k=1}^{K} \tilde{y}_k \bigg( \phi(S) - \phi(\alpha_k) \bigg) \\
    &+ \log \bigg( \frac{\Gamma (\sum_{k=1}^{K} \tilde{\alpha}_{k})}{\Gamma(K) \prod_{k=1}^{K} \Gamma(\tilde{\alpha}_{k})} \bigg) \\
    &+ \sum_{k=1}^{K}(\tilde{\alpha}_{k} - 1)\left[ \psi(\tilde{\alpha}_{k} - \psi(\tilde{S}))\right],
\end{align}
where $\tilde{y}_{k}$ is the pseudo-label obtained from evidential probability, $\tilde{\alpha}_{k} = \tilde{y}_{k} + (1-\tilde{y}_k)\alpha_k$ is adjusted Dirichlet parameters after removing the correct evidence, $\alpha_k=\phi(u_{\theta}(f_{\theta}(x)) + 1$, and $\tilde{S}=\sum_{k=1}^{K}\tilde{\alpha}_k$

Finally, the overall target adaptation loss for E-MAPU can be accomplished by jointly optimizing the above loss as follows,
\begin{align}
    \mathcal{L}_{trg} = \mathcal{L}_{evd} + \beta \mathcal{L}_{mapu}^{T},
\end{align}
where the hyperparameter $\beta$ controls the relative weight between these two loss components.

\subsubsection{Integration with other source-free domain adaptation}
The proposed feature imputation component in MAPU exhibits versatility and can be seamlessly integrated with existing source-free domain adaptation methods during both source pre-training and target adaptation stages.
Specifically, during target adaptation, the objective is to optimize the target encoder $f_\theta$ by balancing the temporal imputation loss and the generic SFDA loss. This optimization achieves both temporal consistency and adaptation on the target domain. The loss function can be formalized as:
\begin{align}
\min_{f_\theta} \mathcal{L}_{trg} = \mathcal{L}_{sfda} + \beta \mathcal{L}_{mapu}^T.
\end{align}

Here, $\beta$ is a hyperparameter that controls the relative weight between the temporal imputation task (governed by $\mathcal{L}_{mapu}^T$) and the generic adaptation objective encoded by $\mathcal{L}_{sfda}$. The latter term represents the loss function employed by the chosen SFDA method to adapt the target domain to the source domain.
For instance, in our MAPU implementation, we combine the temporal imputation loss $\mathcal{L}_{mapu}^T$ with the information maximization loss from SHOT \cite{shot} as the generic source-free loss $\mathcal{L}_{sfda}$ during the adaptation step. Furthermore, the evidential learning component introduced in E-MAPU also demonstrates versatility and can be integrated with various SFDA methods. A more detailed discussion of the versatility of the proposed models is provided in the experiment section.

\section{Experiments}
\subsection{Datasets}
To evaluate the generalizability of our proposed method, we conduct extensive experiments on five real-world datasets encompassing diverse time series applications: machine fault diagnosis, human activity recognition, and sleep stage classification. As shown in Table \ref{tbl:datasets}, these datasets exhibit significant variation across various aspects, resulting in a substantial domain shift across different domains. 

\begin{table}[!tbh]
\centering
\caption{Details of the adopted datasets (C: \#channels, K: \#classes, L: sample length).}
\resizebox{\columnwidth}{!}{
\begin{NiceTabular}{l|ccc|cc}
\toprule
\textbf{Dataset} & \textbf{C}  & \textbf{K} & \textbf{L} & \# training samples & \# testing samples \\ \midrule

UCIHAR &  9 & 6 & 128 & 2300 & 990 \\ 
SSC & 1 & 5 & 3000 & 14280 & 6130 \\ 
MFD & 1 & 3 & 5120 & 7312 & 3604 \\ 
HHAR & 3 & 6 & 128 & 12716 & 5218 \\
WISDM & 3 & 6 & 128 & 1350 & 720 \\
\bottomrule
\end{NiceTabular}
}
\label{tbl:datasets}
\end{table}

\subsubsection{UCIHAR Dataset} 
This dataset is specifically designed for human activity recognition tasks. Data collection involved three sensor types: accelerometers, gyroscopes, and body sensors. Each sensor captures three-dimensional readings, resulting in a total of nine channels per sample with 128 data points each. The data originates from 30 distinct users, with each user treated as a separate domain. To evaluate cross-domain performance, we conducted five experiments where the model was trained on data from one user and tested on data from different users \cite{uciHAR_dataset}.

\subsubsection{Sleep Stage Classification (SSC) Dataset} 
The Sleep Stage Classification (SSC) task aims to categorize Electroencephalogram (EEG) signals into five distinct stages: Wake (W), Non-Rapid Eye Movement (NREM) stages (N1, N2, N3), and Rapid Eye Movement (REM). We evaluate our method on this task using the Sleep-EDF dataset \cite{sleepEDF_dataset}, which contains EEG recordings from 20 healthy subjects. In line with previous studies \cite{attnSleep_paper}, we focus on a single channel (Fpz-Cz) and construct five cross-domain experiments using data from 10 subjects.

\subsubsection{Machine Fault Diagnosis (MFD) Dataset}
This dataset collected by Paderborn University focuses on the application of fault diagnosis, where vibration signals are leveraged to identify various incipient fault types. The data was collected under four distinct operating conditions. Each data sample consists of a single univariate channel with 5120 data points, consistent with previous works \cite{fd_dataset,slarda}. In our experiments, we treat each operating condition as a separate domain and employ five cross-condition scenarios to evaluate the model's domain adaptation performance.

\subsubsection{Heterogeneity Human Activity Recognition (HHAR) Dataset}
This dataset was collected from nine subjects using smartphone and smartwatch sensor readings. To mitigate inter-subject variability, we employed a uniform data collection approach, utilizing the same Samsung smartphone device for all subjects during data acquisition. Three types of sensors are used to record the data, thus each data sample has three multivariate time series, where each channel has 128 data points. For the experiments, we treat each subject's data as a distinct domain. Five cross-domain scenarios were then formulated by randomly selecting subjects from the pool \cite{hhar_dataset}.

\subsubsection{Wireless Sensor Data Mining (WISDM) Dataset}
This dataset also employs accelerometer sensors, collecting data from 36 subjects performing activities identical to those in the UCIHAR dataset. Three types of sensors are used to record the data, resulting in a total of three channels per sample with 128 data points each. However, it presents a potentially greater challenge due to class imbalance issues within individual subject data. Following the approach used for UCIHAR, we treat each subject's data as a distinct domain and formulate five cross-domain scenarios by randomly selecting subjects \cite{wisdm_dataset}.

More details about the datasets are included in Table \ref{tbl:datasets}.

\subsection{Baselines}
To assess the effectiveness of our model, we compare its performance against established UDA approaches that presume access to source data during adaptation. These baseline models are derived from the AdaTime benchmark \cite{adatime}. We further evaluate our method against recent SFDA techniques. To guarantee a fair comparison, all SFDA baselines were re-implemented within our framework, employing the identical backbone network and training configurations. Here's a summarized list of the compared methods:

\subsubsection{Conventional UDA methods}

\begin{itemize}
    \item Deep Domain Confusion (DDC) \cite{ddc}: leverages the Maximum Mean Discrepancy (MMD) distance to align the source and target feature distributions, aiming to minimize domain shift.
    \item Deep Correlation Alignment (DCORAL) \cite{deep_coral}: focuses on aligning the second-order statistics (variances and covariances) of the source and target distributions for effective domain adaptation.
    \item High-order Maximum Mean Discrepancy (HoMM) \cite{HoMM}: extends MMD by incorporating higher-order moments, aiming to capture more complex relationships in the data and improve adaptation performance.
    \item Minimum Discrepancy Estimation for Deep Domain Adaptation (MMDA) \cite{MMDA}: combines MMD and correlation alignment with entropy minimization to comprehensively address domain shift issues.
    \item Domain-Adversarial Training of Neural Networks (DANN) \cite{DANN}: employs a gradient reversal layer to train a domain discriminator network adversarially against an encoder network. This approach encourages the encoder to learn domain-agnostic representations.
    \item Conditional Domain Adversarial Network (CDAN) \cite{CDAN}: introduces conditional adversarial alignment, integrating task-specific knowledge with the features during the alignment step to improve adaptation across different domains.
    \item Convolutional deep adaptation for time series (CoDATS) \cite{codats}: leverages adversarial training with weak supervision to enhance the adaptation performance specifically on time series data.
    \item Adversarial spectral kernel matching (AdvSKM)\cite{dskn}: introduces adversarial spectral kernel matching to address the challenges of non-stationarity and non-monotonicity inherent in time series data.
\end{itemize}

\subsubsection{Source-free DA methods}

\begin{itemize}
    \item Source Hypothesis Transfer (SHOT) \cite{shot}: minimizes information maximization loss with self-supervised pseudo-labels. This process aims to identify target features compatible with the transferred source hypothesis, facilitating adaptation without source labels.
    \item Exploiting the intrinsic neighborhood structure (NRC) \cite{nrc}: captures the intrinsic structure of the target data. It achieves this by forming clear clusters and encouraging label consistency among data points with high local affinity, effectively addressing the lack of labeled target data.
    \item Attracting and dispersing (AaD) \cite{aad}: optimizes a prediction consistency objective, encouraging local neighborhood features in the feature space to have similar predictions. This strategy aims to leverage the inherent structure within the unlabeled target data.
\end{itemize}

\subsection{Implementation Details}
\subsubsection{Encoder Architecture}
Consistent with existing works \cite{adatime,ts_tcc,ca_tcc}, we leverage a 1-dimensional convolutional neural network (CNN) encoder architecture for our model. This encoder consists of three convolutional layers with filter sizes of 64, 128, and 128, respectively. Each convolutional layer is followed by a rectified linear unit (ReLU) activation function and batch normalization for improved learning performance.

\subsubsection{Network Parameters}
Our experiments employ a consistent temporal masking strategy across all datasets. A masking ratio of 1/8 is applied, meaning that 12.5\% of the time series data points are randomly masked during training. For imputation, we utilize a single-layer recurrent neural network (RNN) with a hidden dimension of 128 for all datasets.

\subsubsection{Unified Training Scheme}
To ensure a fair and rigorous comparison with existing source-free baseline methods \cite{shot, nrc, aad}, we adopted their established implementations. We further ensured consistency by employing the identical backbone network and training procedures used in our proposed method.

Consistent with the AdaTime framework \cite{adatime}, we train all models for 40 epochs with a batch size of 32. The learning rate is set to 1e-3 for UCIHAR and 3e-3 for both SSC and MFD datasets. We built our model using Pytorch and trained it on an NVIDIA GeForce RTX 2080Ti GPU. To ensure robust evaluation under potential data imbalance, we employ the macro F1-score (MF1) metric \cite{adatime}. We report the mean and standard deviation of the metric across three consecutive runs for each cross-domain scenario.

\begin{table*}[h]
\centering
\caption{Detailed results of the five UCIHAR cross-domain scenarios in terms of MF1 score.}
\begin{NiceTabular}{@{}l|c|ccccc|c@{}} 
\toprule 
Algorithm & SF & 2$\rightarrow$11 & 12$\rightarrow$16 & 9$\rightarrow$18 & 6$\rightarrow$23 & 7$\rightarrow$13 & AVG\\ \midrule
DDC & \xmark & 60.0$\pm$13.32 & 66.77$\pm$8.46 & 61.41$\pm$5.80 & 88.55$\pm$1.42 & 77.29$\pm$2.11 &   75.67 \\
DCoral &\xmark &  67.2$\pm$13.67 & 64.58$\pm$8.72 & 54.38$\pm$9.69 & 89.66$\pm$2.54 & 90.46$\pm$2.96 &   77.71 \\
HoMM &\xmark &  83.54$\pm$2.99 & 63.45$\pm$2.07 & 71.25$\pm$4.42 & 94.97$\pm$2.49 & 91.41$\pm$1.33 &   84.10 \\
MMDA &\xmark &  72.91$\pm$2.78 & \textbf{74.64$\pm$2.88}& 62.62$\pm$2.63 & 91.14$\pm$0.46 & 90.61$\pm$2.00 &  81.40  \\
DANN &\xmark &  98.09$\pm$1.68 & 62.08$\pm$1.69 & 70.7$\pm$11.36 & 85.6$\pm$15.71 & 93.33$\pm$0.00 &   84.97 \\
CDAN &\xmark &  \underline{98.19$\pm$1.57} & 61.20$\pm$3.27 & 71.3$\pm$14.64 & {96.73$\pm$0.00} & 93.33$\pm$0.00  &   86.79 \\
CoDATS &\xmark &  86.65$\pm$4.28 & 61.03$\pm$2.33 & 80.51$\pm$8.47 & 92.08$\pm$4.39 & 92.61$\pm$0.51  &   85.47 \\
AdvSKM  &\xmark &  65.74$\pm$2.69 & 60.52$\pm$1.99 & 53.25$\pm$5.19 & 79.63$\pm$8.52 & 88.89$\pm$3.12 &   74.67 \\ 

\midrule

SHOT & \cmark  &\textbf{100.0$\pm$0.00}   & 70.76$\pm$6.22  & 70.19$\pm$8.99 & \textbf{98.91$\pm$1.89}  & 93.01$\pm$0.57	& 86.57 \\ 
NRC & \cmark  & 97.02$\pm$2.82  &72.18$\pm$0.59  & 63.10$\pm$4.84  & 96.41$\pm$1.33 & 89.13$\pm$0.54  & 83.57   \\ 
AaD  & \cmark & 98.51$\pm$2.58 & 66.15$\pm$6.15  & 68.33$\pm$11.9 & \underline{98.07$\pm$1.71} & 89.41$\pm$2.86  & 84.09 \\ 

\midrule

\textbf{MAPU} & \cmark  & \textbf{100.0$\pm$0.00}  &  67.96$\pm$4.62  & \underline{82.77$\pm$2.54}  & 97.82$\pm$1.89  & \textbf{99.29$\pm$1.22} & \underline{89.57 } \\ 
\textbf{E-MAPU} & \cmark & \textbf{100.0$\pm$0.00} & \underline{72.85$\pm$2.13} & \textbf{88.37$\pm$8.47} & 97.82$\pm$1.54 & \underline{98.59$\pm$0.50} & \textbf{91.53} \\

\bottomrule
\end{NiceTabular}
\label{table:har}
\end{table*}



\begin{table*}[]
\centering
\caption{Detailed results of the five SSC cross-domain scenarios in terms of MF1 score.}
\begin{NiceTabular}{l|c|ccccc|c} 
\toprule 
Algorithm & SF & 16$\rightarrow$1 & 9$\rightarrow$14 & 12$\rightarrow$5 & 7$\rightarrow$18 & 0$\rightarrow$11 & AVG\\ \midrule
DDC & \xmark & 55.47$\pm$1.72 & 63.57$\pm$1.43 & 55.43$\pm$2.75 & 67.46$\pm$1.45 & \underline{54.17$\pm$1.79} & 59.22 \\
DCoral &  \xmark& 55.50$\pm$1.74 & 63.50$\pm$1.36 & 55.35$\pm$2.64 & 67.49$\pm$1.50 & 53.76$\pm$1.89 & 59.12 \\
HoMM &\xmark& 55.51$\pm$1.79 & 63.49$\pm$1.14 & 55.46$\pm$2.71 & 67.50$\pm$1.50 & 53.37$\pm$2.47 & 59.06 \\
MMDA &\xmark& 62.92$\pm$0.96 & 71.04$\pm$2.39 & \textbf{65.11$\pm$1.08} & \underline{70.95$\pm$0.82} & 43.23$\pm$4.31 & 62.79 \\
DANN &\xmark& 58.68$\pm$3.29 & 64.29$\pm$1.08 & 64.65$\pm$1.83 & 69.54$\pm$3.00 & 44.13$\pm$5.84 & 60.26 \\
CDAN &\xmark& 59.65$\pm$4.96 & 64.18$\pm$6.37 & 64.43$\pm$1.17 & 67.61$\pm$3.55 & 39.38$\pm$3.28 & 59.04 \\
CoDATS &\xmark& \underline{63.84$\pm$3.36} & 63.51$\pm$6.92 & 52.54$\pm$5.94 & 66.06$\pm$2.48 & 46.28$\pm$5.99 & 58.44 \\
AdvSKM &\xmark& 57.83$\pm$1.42 & 64.76$\pm$3.00 & 55.73$\pm$1.42 & 67.58$\pm$3.64 & \textbf{55.19$\pm$4.19} & 60.21 \\ 

\midrule

SHOT & \cmark & 59.07$\pm$2.14 & 69.93$\pm$0.46 & 62.11$\pm$1.62 & 69.74$\pm$1.22 & 50.78$\pm$1.90 & 62.33 \\ 
NRC & \cmark & 52.09$\pm$1.89 & 58.52$\pm$0.66 & 59.87$\pm$2.48 & 66.18$\pm$0.25 & \underline{47.55$\pm$1.72} & 56.84 \\ 
AaD & \cmark & 57.04$\pm$2.03 & 65.27$\pm$1.69 & 61.84$\pm$1.74 & 67.35$\pm$1.48 & 44.04$\pm$2.18 & 59.11 \\ 

\midrule

\textbf{MAPU} & \cmark & \textbf{63.85$\pm$4.63} & \textbf{74.73$\pm$0.64} & 64.08$\pm$2.21 & \textbf{74.21$\pm$0.58} & 43.36$\pm$5.49 & \underline{64.05} \\
\textbf{E-MAPU} & \cmark & 62.22$\pm$0.33 & \underline{73.66$\pm$0.43} & \underline{64.88$\pm$2.64} & 70.59$\pm$1.35 & 51.24$\pm$4.84 & \textbf{64.52} \\

\bottomrule
\end{NiceTabular}
\label{table:eeg}
\end{table*}



\begin{table*}[!ht]
    \centering
    \caption{Detailed results of the five MFD cross-domain scenarios in terms of MF1 score.}
    \begin{NiceTabular}{@{}l|c|ccccc|c@{}} 
    \midrule
    Algorithm &SF & 0$\rightarrow$1 & 1$\rightarrow$0 & 1$\rightarrow$2 & 2$\rightarrow$3 & 3$\rightarrow$1 & AVG \\ \toprule
    DDC & \xmark & 74.50$\pm$5.56 & 48.91$\pm$6.24 & 89.34$\pm$2.16 & 96.34$\pm$3.07 & \textbf{100.0$\pm$0.00} & 81.82  \\ 
    DCoral & \xmark & 79.03$\pm$8.83 & 40.83$\pm$5.01 & 82.71$\pm$0.76 & 98.01$\pm$0.67 & 97.73$\pm$3.93 & 79.66  \\
    HoMM & \xmark & 80.80$\pm$2.46 & 42.31$\pm$5.90 & 84.28$\pm$1.32 & 98.61$\pm$0.08 & 96.28$\pm$6.45 & 80.46  \\
    MMDA & \xmark & 82.44$\pm$4.47 & 49.35$\pm$5.02 & \textbf{94.07$\pm$2.72} & \textbf{100.0$\pm$0.00} & \textbf{100.0$\pm$0.00} & 85.17  \\ 
    DANN & \xmark & 83.44$\pm$1.72 & 51.52$\pm$0.38 & 84.19$\pm$2.10 & \underline{99.95$\pm$0.09} & \textbf{100.0$\pm$0.00} & 83.82  \\ 
    CDAN & \xmark & 84.97$\pm$0.62 & 52.39$\pm$0.49 & 85.96$\pm$0.90 & 99.70$\pm$0.45 & \textbf{100.0$\pm$0.00} & 84.60 \\ 
    CoDATS & \xmark & 67.42$\pm$13.3 & 49.92$\pm$13.7 & 89.05$\pm$4.73 & 99.21$\pm$0.79 & 99.92$\pm$0.14 & 81.10  \\ 
    AdvSKM & \xmark & 76.64$\pm$4.82 & 43.81$\pm$6.29 & 83.10$\pm$2.19 & 98.85$\pm$0.93 & \textbf{100.0$\pm$0.00} & 80.48  \\ 
    
    \midrule
    
    SHOT & \cmark& 41.99$\pm$2.78 & 57.00$\pm$0.09 & 80.70$\pm$1.49 & 99.48$\pm$0.31 & 99.95$\pm$0.05 & 75.82  \\ 
    NRC & \cmark& 73.99$\pm$1.36 & 74.88$\pm$8.81 & 69.23$\pm$0.75 & 78.04$\pm$11.3 & 71.48$\pm$4.59 & 73.52 \\ 
    AaD & \cmark & 71.72$\pm$3.96 & \underline{75.33$\pm$4.65} & 78.31$\pm$2.26 & 90.07$\pm$7.02 & 87.45$\pm$11.7 & 80.58  \\
    
    \midrule 
    
    \textbf{MAPU}  & \cmark&  \textbf{99.43$\pm$0.51} & \underline{77.42$\pm$0.16} & 85.78$\pm$7.38 & 99.67$\pm$0.50 & \underline{99.97$\pm$0.05} & \underline{92.45}\\
    \textbf{E-MAPU}  & \cmark&  \underline{98.83$\pm$0.61} & \textbf{89.43$\pm$0.40} & \underline{92.05$\pm$0.74} & 99.89$\pm$0.10 & 99.95$\pm$0.04 & \textbf{96.03}\\

    \bottomrule
    
    \end{NiceTabular}
    \label{table:fd}
\end{table*}



\subsection{Comparative Experiments}
In this section, we rigorously test our approach against state-of-the-art methods in various time series applications. To assess the efficacy of our approach, we evaluate its performance on five different time series datasets, namely, UCIHAR, SSC, MFD, HHAR, and WISDM. Tables \ref{table:har}, \ref{table:eeg}, \ref{table:fd}, \ref{table:hhar}, and \ref{table:wisdm} present results for five cross-domain scenarios in each dataset, as well as an average performance across all scenarios. The algorithms are divided into two groups: the traditional UDA methods are marked with \xmark,  while the source-free methods are marked with \cmark.

\subsubsection{Quantative Results on UCIHAR Dataset}
Table \ref{table:har} presents the performance of our proposed methods (MAPU and E-MAPU) and prior works in five cross-subject scenarios on the UCIHAR dataset. MAPU achieves superior performance in two out of the five scenarios, with an overall accuracy of 89.57\%. This surpasses the second-best source-free baseline by 3\%. Interestingly, these source-free methods (i.e., SHOT, NRC, and AaD) perform competitively with conventional UDA methods that leverage source data during training. This can be attributed to their two-stage training scheme (pertaining and adaptation) that focuses on optimizing the target model for the target domain without relying on source performance \cite{cpga}. Furthermore, MAPU, with its built-in temporal adaptation capability, outperforms all conventional UDA methods, surpassing the best-performing method (e.g., CDAN) by 2.78\%. Notably, E-MAPU, which incorporates evidential uncertainty learning, achieves an even better overall performance of 91.53\%. This suggests that reducing evidential entropy leads to improved alignment between the target and source domain features.

\begin{table*}[h]
\centering
\caption{Detailed results of the five HHAR cross-domain scenarios in terms of MF1 score.}
\begin{NiceTabular}{@{}l|c|ccccc|c@{}} 
\toprule 
Algorithm & SF & 0$\rightarrow$6 & 1$\rightarrow$6 & 2$\rightarrow$7 & 3$\rightarrow$8 & 4$\rightarrow$5 & AVG\\ \midrule
DDC & \xmark & 53.33$\pm$14.28 & 80.89$\pm$5.83 & 44.42$\pm$5.09 & 76.47$\pm$5.36 & 81.24$\pm$1.83 & 67.27 \\
DCoral &\xmark &  54.19$\pm$11.19 &  86.72$\pm$2.26 &  46.22$\pm$4.31 &  72.26$\pm$10.54 &  87.15$\pm$2.84 &  69.31 \\
HoMM &\xmark &  50.88$\pm$14.30 &  89.13$\pm$1.00 &  42.51$\pm$1.41 &  80.72$\pm$1.84 &  90.73$\pm$1.73 &  70.79 \\
MMDA &\xmark &  49.56$\pm$2.42 &  92.81$\pm$1.41 &  56.27$\pm$5.63 &  97.33$\pm$1.06 &  97.04$\pm$0.63 &  78.60  \\
DANN &\xmark &  45.01$\pm$1.12 &  93.64$\pm$0.86 &  50.71$\pm$6.39 &  96.80$\pm$0.96 &  96.99$\pm$0.95 &  76.63 \\
CDAN &\xmark &  45.35$\pm$0.59 &  \underline{94.12$\pm$0.33} &  50.33$\pm$6.63 &  97.37$\pm$0.48 &  87.77$\pm$14.94 &  74.99 \\
CoDATS &\xmark &  38.05$\pm$8.70 &  92.46$\pm$0.70 &  50.88$\pm$3.83 &  94.43$\pm$1.52 &  86.50$\pm$1.99 &  72.46 \\
AdvSKM  &\xmark &  58.50$\pm$3.56 &  64.22$\pm$12.46 &  39.07$\pm$3.03 &  78.87$\pm$0.28 &  80.71$\pm$3.90 &  64.27 \\ 

\midrule

SHOT & \cmark  & 64.16$\pm$0.24 & 91.77$\pm$0.63 & \textbf{65.81$\pm$0.37} & \underline{98.41$\pm$1.09} & 96.99$\pm$0.22 & 83.43 \\ 
NRC & \cmark  & 68.02$\pm$0.17 & 90.62$\pm$1.14 & 45.42$\pm$0.56 & 81.30$\pm$1.56 & 94.94$\pm$1.36 & 76.06   \\ 
AaD  & \cmark & \underline{72.31$\pm$15.94} & 91.06$\pm$1.18 & 52.91$\pm$15.47 & 85.51$\pm$5.92 & 96.37$\pm$0.38 & 79.63 \\ 

\midrule

\textbf{MAPU} & \cmark  & 65.49$\pm$2.12 & \textbf{94.20$\pm$0.10} & \underline{64.67$\pm$0.29} & \textbf{99.27$\pm$0.01} & \underline{97.27$\pm$0.16} & \underline{84.18} \\ 
\textbf{E-MAPU} & \cmark & \textbf{79.43$\pm$13.34} & 94.05$\pm$0.01 & 57.88$\pm$7.75 & \textbf{99.27$\pm$0.01} & \textbf{97.70$\pm$0.09} & \textbf{85.67} \\

\bottomrule
\end{NiceTabular}
\label{table:hhar}
\end{table*}
\begin{table*}[h]
\centering
\caption{Detailed results of the five WISDM cross-domain scenarios in terms of MF1 score.}
\begin{NiceTabular}{@{}l|c|ccccc|c@{}} 
\toprule 
Algorithm & SF & 6$\rightarrow$19 & 2$\rightarrow$11 & 33$\rightarrow$12 & 5$\rightarrow$26 & 28$\rightarrow$4 & AVG\\ \midrule
DDC & \xmark & 64.62$\pm$7.54 & 69.35$\pm$14.44 & 43.19$\pm$0.11 & 29.30$\pm$0.86 & 74.31$\pm$4.30 & 56.15 \\
DCoral &\xmark &  62.58$\pm$4.27 & 71.35$\pm$11.93 & 43.81$\pm$0.13 & 29.32$\pm$0.68 & 73.78$\pm$4.49 & 56.17 \\
HoMM &\xmark &  61.37$\pm$10.29 & 55.50$\pm$15.90 & \underline{62.12$\pm$1.36} & 30.30$\pm$1.54 & 68.80$\pm$3.45 & 55.62 \\
MMDA &\xmark &  56.00$\pm$0.24 & 59.40$\pm$8.11 & 53.23$\pm$5.82 & 27.65$\pm$2.51 & 79.67$\pm$4.42 & 55.19  \\
DANN &\xmark &  51.96$\pm$18.35 & 66.52$\pm$4.03 & 54.19$\pm$7.53 & 30.00$\pm$1.05 & \textbf{88.79$\pm$0.80} & 58.29 \\
CDAN &\xmark &  47.19$\pm$6.65 & 44.06$\pm$2.34 & 58.80$\pm$7.39 & 30.32$\pm$4.06 & \underline{86.10$\pm$7.83} & 53.29 \\
CoDATS &\xmark &  62.67$\pm$13.40 & 57.08$\pm$7.61 & \textbf{67.76$\pm$7.94} & 35.75$\pm$5.20 & 84.87$\pm$4.78 & \underline{61.63} \\
AdvSKM  &\xmark &  60.17$\pm$7.06 & 73.49$\pm$2.29 & 46.03$\pm$1.58 & 30.15$\pm$1.03 & 70.34$\pm$2.66 & 56.04 \\ 

\midrule

SHOT & \cmark  & \textbf{72.41$\pm$2.75} & 69.01$\pm$7.5 & 55.81$\pm$2.10 & 29.78$\pm$3.05 & 65.07$\pm$6.92 & 58.42 \\ 
NRC & \cmark  & 68.45$\pm$6.54 & \textbf{81.76$\pm$9.86} & 45.0$\pm$0.74 & 28.73$\pm$3.52 & 61.26$\pm$2.74 & 57.04  \\ 
AaD  & \cmark & 60.63$\pm$0.00 & 68.48$\pm$1.73 & 42.51$\pm$4.27 & \underline{37.96$\pm$3.65} & 76.79$\pm$1.03 & 57.28 \\ 

\midrule

\textbf{MAPU} & \cmark  & \underline{69.85$\pm$4.41} & \underline{78.06$\pm$1.24} & 53.06$\pm$9.66 & 31.72$\pm$1.11 & 74.36$\pm$3.58 & 61.41 \\ 
\textbf{E-MAPU} & \cmark & 69.58$\pm$11.82 & 72.68$\pm$3.35 & 60.79$\pm$8.20 & \textbf{40.51$\pm$1.44} & 76.66$\pm$1.47 & \textbf{64.04} \\

\bottomrule
\end{NiceTabular}
\label{table:wisdm}
\end{table*}

\subsubsection{Quantative Results on SSC Dataset}
Table \ref{table:eeg} showcases the superior performance of our proposed methods over other baselines on the sleep stage classification task. MAPU achieves the best results in three out of the five cross-domain scenarios, reaching an overall performance of 64.05\%. This surpasses the best source-free method (e.g., SHOT) by 1.72\% and the best conventional UDA method by 1.27\%. Notably, source-free methods like NRC and AaD, which rely on feature clustering, perform poorly on the sleep stage classification dataset due to the inherent class imbalance. However, MAPU, with its temporal adaptation capability, effectively handles such imbalance, outperforming all source-free methods by a maximum improvement of 4.8\% (scenario 16 $\rightarrow$ 1). Furthermore, E-MAPU, incorporating evidential uncertainty, further improves upon MAPU, reaching an overall performance of 64.52\%. This suggests the effectiveness of the well-calibrated model with evidential uncertainty estimations in enhancing domain adaptation performance.

\subsubsection{Quantative Results on MFD Dataset}
Table \ref{table:fd} highlights the superior performance of MAPU on the machine fault diagnosis task. MAPU achieves an outstanding average performance of 92.45\%, exceeding the second-best baseline method by a significant margin of 7.85\%. Notably, MAPU demonstrates exceptional performance in challenging transfer scenarios (0 $\rightarrow$ 1 and 1 $\rightarrow$ 0), achieving a 14.46\% improvement in the latter. While remaining competitive in easier transfer tasks (2 $\rightarrow$ 3 and 3 $\rightarrow$ 1), MAPU outperforms all baselines across all cross-domains. Compared to source-free methods, MAPU surpasses the second-best (AaD) by a substantial margin of 11.87\%. Furthermore, E-MAPU significantly advances the state-of-the-art, pushing the average F1-score from 92.45\% to 96.03\%. This emphasizes the effectiveness of the evidential entropy optimization strategy in mitigating overconfidence and reducing discrepancies between source and target domain distributions.

It is worth noting that the performance improvement of our method is relatively large in the MFD dataset compared to other datasets. The larger performance gains observed on the MFD dataset compared to others can be attributed to two key factors. First, its sequences are the longest among all other datasets, making temporal adaptation more critical. Second, unlike other datasets, the dataset has only three classes, meaning misclassifications of a single class have a more substantial impact on overall performance.

\subsubsection{Quantative Results on HHAR Dataset}
Table \ref{table:hhar} summarizes the performance of our proposed methods (MAPU and E-MAPU) alongside prior works on the HHAR dataset across five cross-subject scenarios. MAPU demonstrates superior performance in two scenarios, achieving an overall accuracy of 84.18\%. This surpasses the best traditional domain adaptation method by over 5\% and the second-best source-free SHOT method by nearly 1\%. Notably, E-MAPU further improves upon MAPU by almost 1.5\%, achieving the best performance in three scenarios and an overall optimal f1-score of 85.67\%. Significantly, E-MAPU with its built-in evidential uncertainty estimation capability, outperforms all conventional UDA methods. It surpasses the best-performing method (e.g., MMDA) by nearly 7\%. This improvement can be attributed to the use of evidential deep learning, which leads to a better-calibrated model and consequently enhances adaptation performance.

\subsubsection{Quantative Results on WISDM Dataset}
Table \ref{table:wisdm} summarizes the performance of MAPU and E-MAPU compared to existing methods across five cross-domain scenarios on the WISDM dataset. MAPU consistently outperforms all source-free methods, surpassing the second-best SHOT method by nearly 3\% and achieving overall better performance than traditional domain adaptation, which performed comparably to the optimal traditional method. This demonstrates the significant improvement in adaptation facilitated by temporal adaptation strategies within MAPU. E-MAPU builds upon this success by incorporating evidential deep learning strategies. Consequently, it outperforms all compared baseline methods, achieving a state-of-the-art classification performance of 64.04\%. It is important to note that the WISDM dataset presents additional challenges due to class imbalance issues within individual subjects' data.  Despite this, our proposed methods, leveraging both temporal adaptation and uncertainty estimation strategies, significantly enhance the reliability of the adaptation model.

\begin{figure*}[t]
\centering
\begin{tabular}{c}
\subfigure[]{\includegraphics[width=0.3\linewidth]{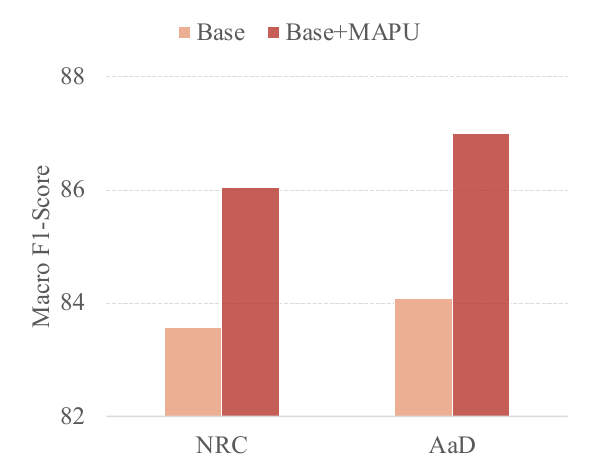} \label{fig_mapu_others_a}}
\subfigure[]{\includegraphics[width=0.3\linewidth]{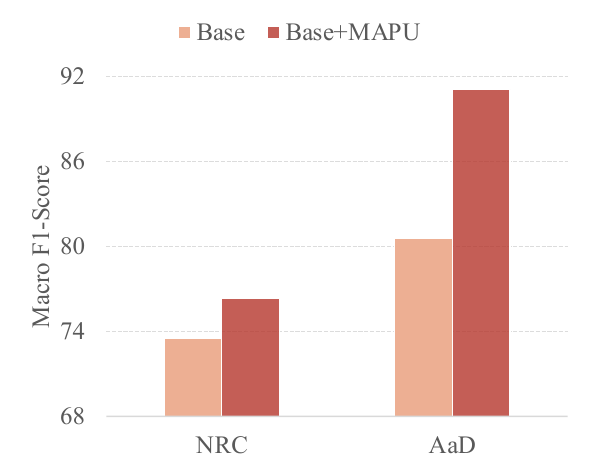} \label{fig_mapu_others_b}}
\subfigure[]{\includegraphics[width=0.3\linewidth]{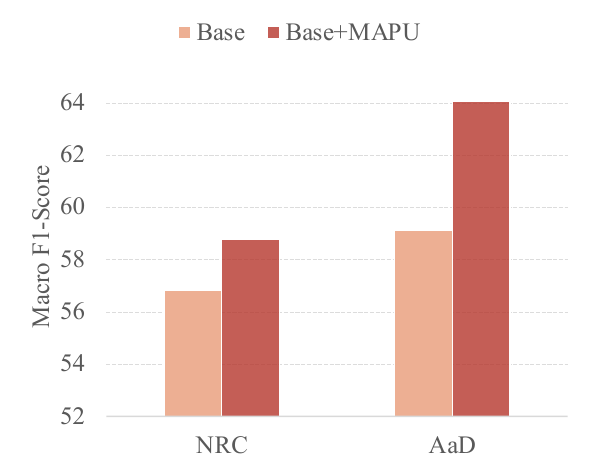} \label{fig_mapu_others_c}}
\\
\subfigure[]{\includegraphics[width=0.3\linewidth]{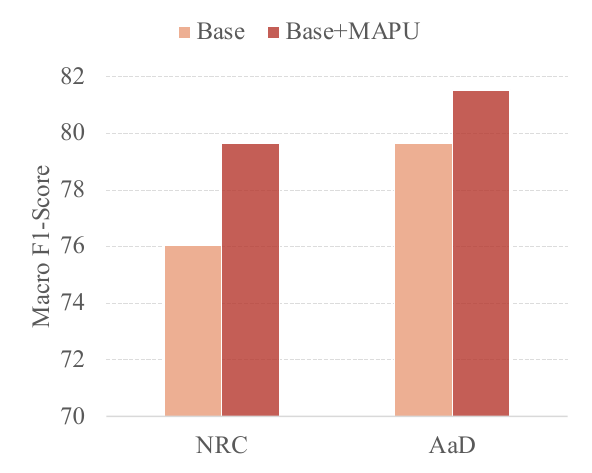} \label{fig_mapu_others_d}}
\subfigure[]{\includegraphics[width=0.3\linewidth]{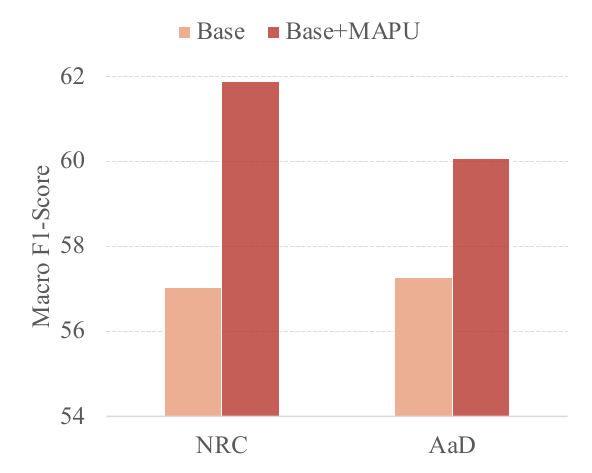} \label{fig_mapu_others_e}}

\end{tabular}
\caption{Intergrating temporal imputation with existing SFDA methods among the three datasets. (a) UCIHAR dataset. (b) MFD dataset. (c) SSC dataset. (d) HHAR dataset. (e) WISDM dataset.}
\label{fig_mapu_others}
\end{figure*}

\section{Model Analysis and Discussion}
\subsection{Ablation Study on Auxiliary Tasks}
To evaluate the effectiveness of our proposed temporal imputation auxiliary task, we compared it against alternative approaches commonly used in SFDA, including rotation prediction \cite{SHOT++} and jigsaw puzzle reconstruction \cite{sl-sfda} on the five datasets. Three different SFDA backbones (SHOT, NRC, and AaD) were employed with each auxiliary task to minimize bias towards a specific SFDA method. Table \ref{tab:aux} summarizes the average performance across five cross-domain scenarios for each dataset. The results consistently demonstrate that our temporal imputation task outperforms the other auxiliary tasks on all datasets, even when combined with different SFDA backbones. Conversely, the baseline tasks (rotation prediction and jigsaw puzzle) show limited improvement and, in many cases across datasets, even hinder performance. This highlights the limitations of these tasks for time series data and emphasizes the importance of incorporating temporal dynamics into the adaptation process, as demonstrated by the superior performance of our imputation approach.
\begin{table}[]
\centering
\caption{Comparing the temporal imputation task with conventional auxiliary tasks for time series adaptation.}
\begin{NiceTabular}{l|ccccc}
\toprule
Task & UCIHAR & SSC & MFD & HHAR & WISDM \\ \midrule
SHOT & 86.57 & 62.33 & 75.82  & 83.43 & 58.42 \\ 
SHOT+   Rotation & 86.78 & 60.33 & 84.98 & 83.85 & 58.44 \\
SHOT + Jigsaw & 87.83 & 62.11 & 85.74 & 82.76 & 57.42 \\
\textbf{SHOT + Temporal} & \textbf{89.57} & \textbf{64.05} & \textbf{92.45} & \textbf{84.18} & \textbf{61.41} \\

\midrule
NRC              & 83.57 & 56.84 & 73.52 & 76.06 & 57.04 \\
NRC+ Rotation    & 71.62 & 56.75 & 72.02 & \textbf{79.98} & 57.51 \\
NRC + Jigsaw     & 70.58 & 56.91 & 74.68 & 79.49 & 56.60 \\
\textbf{NRC + Temporal}       & \textbf{86.05} &\textbf{58.78} & \textbf{76.34} & 79.63 & \textbf{61.90} \\\midrule
AaD              & 84.09 & 59.11 & 80.58 & 79.63 & 57.28 \\
AaD +   Rotation & 71.52 & 59.00 & 84.18 & 81.22 & 57.30 \\
AaD + Jigsaw     & 83.72 & 59.17 & 85.31 & 79.90 & 55.45 \\
\textbf{AaD + Temporal}       & \textbf{87.00} &\textbf{64.05} & \textbf{91.11} & \textbf{81.52} & \textbf{60.09} \\
\bottomrule
\end{NiceTabular}
\label{tab:aux}
\end{table}

\subsection{Model Versatility Analysis}
\subsubsection{Integrating MAPU with Other SFDA Methods}
This section investigates the efficacy of incorporating temporal information into established SFDA methods. We evaluate the performance of three such methods (SHOT, NRC, and AaD) when combined with our proposed temporal imputation task on the five datasets. Fig. \ref{fig_mapu_others} presents the average performance across five cross-domain scenarios for each dataset. The results consistently demonstrate significant performance improvements across all datasets upon integration with our temporal imputation task. For instance, on the UCIHAR dataset, both NRC and AaD methods experience a notable 3\% boost in performance. These improvements are consistent across other datasets, highlighting the effectiveness of our approach in enhancing the temporal adaptation capabilities of existing SFDA methods, which were primarily designed for visual applications.

\subsubsection{Integrating E-MAPU with Other SFDA Methods}
We further explore the effectiveness of incorporating evidential uncertainty estimation into existing SFDA methods. Similar to the previous section, we evaluate the performance of three methods (SHOT, NRC, and AaD) when integrated with our proposed evidential learning components on the five datasets. Table \ref{tab:mapu_plus_plus_others} presents the average performance across five cross-domain scenarios for each dataset. The results consistently demonstrate significant performance improvements across various SFDA methods when combined with our evidential uncertainty approach. Specifically, on the UCIHAR dataset, both NRC and AaD methods achieve nearly a 4\% performance boost. On the HHAR and WISDM datasets, almost all of the integrated approaches achieve greater than 2\% performance improvement. For the SSC dataset, NRC and AaD experience improvements of approximately 6\% and 3\%, respectively. Notably, on the MFD dataset, both NRC and SHOT methods exhibit average performance gains exceeding 10\%, while AaD achieves a performance boost of almost 6\%. These improvements highlight the effectiveness of our calibrated adaptation using evidential uncertainty in enhancing the reliability and adaptation capabilities of existing SFDA methods.

\begin{table}[h]
\centering
\caption{Intergrating evidential deep learning with existing SFDA methods among the five datasets.}
\begin{NiceTabular}{l|ccccc}
\toprule
Variants & UCIHAR & SSC & MFD & HHAR & WISDM \\ \midrule
SHOT & 86.57 & 62.33 & 75.82 & 83.43 & 58.42  \\ 
\textbf{E-SHOT} & \textbf{87.74} & \textbf{62.90} & \textbf{90.37} & \textbf{83.64} & \textbf{60.49} \\

\midrule
NRC              & 83.57 & 56.84 & 73.52 & 76.06 & 57.04 \\
\textbf{E-NRC}       & \textbf{87.61} &\textbf{ 62.91} & \textbf{88.02} & \textbf{78.09} & \textbf{60.12} \\\midrule
AaD              & 84.09 & 59.11 & 80.58 & 79.63 & 57.28 \\
\textbf{E-AaD}       & \textbf{88.01} &\textbf{ 62.22} & \textbf{86.21} & \textbf{83.41} & \textbf{60.21} \\
\bottomrule
\end{NiceTabular}
\label{tab:mapu_plus_plus_others}
\end{table}

\subsection{Parameter Analysis}
\subsubsection{Sensitivity Analysis}
\begin{figure}[t]
\centering
\begin{tabular}{c}
\subfigure[]{\includegraphics[width=0.5\linewidth]{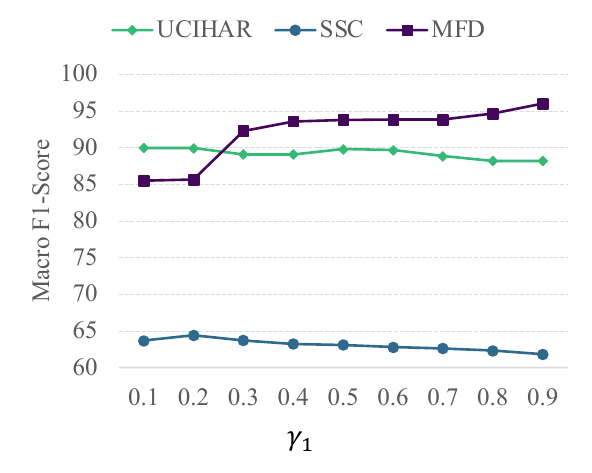} \label{fig_param_sensitivity_a}}
\subfigure[]{\includegraphics[width=0.5\linewidth]{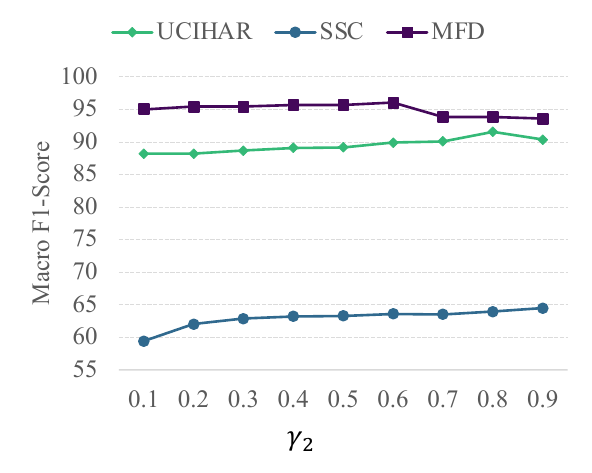} \label{fig_param_sensitivity_b}}
\\
\subfigure[]{\includegraphics[width=0.5\linewidth]{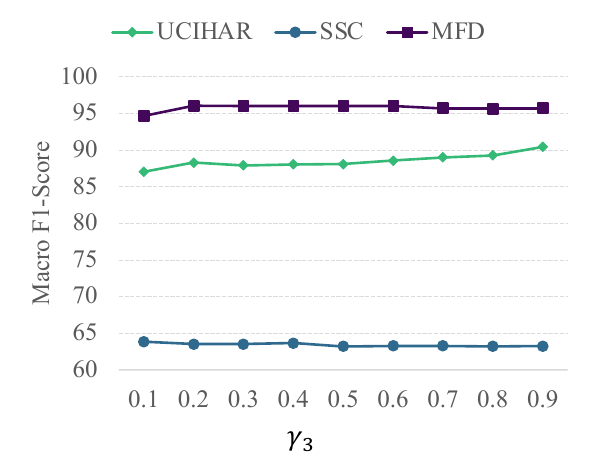} \label{fig_param_sensitivity_c}}
\subfigure[]{\includegraphics[width=0.5\linewidth]{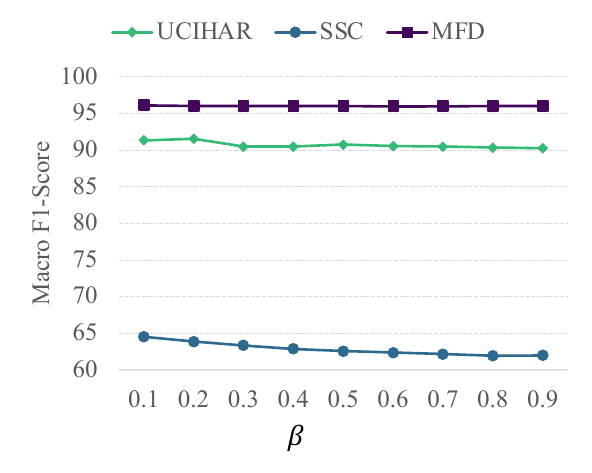} \label{fig_param_sensitivity_d}}

\end{tabular}
\caption{Analysis of adaptation performance with varying relative weight for the model parameters. (a) Evidential entropy. (b) Evidential diversity. (c) Evidential cross-entropy. (d) Temporal imputation.}
\label{fig_param_sensitivity}
\end{figure}

Our method incorporates four key parameters: evidential entropy ($\gamma_1$), evidential diversity ($\gamma_2$), evidential cross-entropy ($\gamma_3$), and the temporal imputation component ($\beta$). To understand their impact on performance, we conducted a sensitivity analysis on the UCIHAR, SSC, and MFD datasets. The parameter range was set from $0.1$ to $0.9$ with increments of $0.1$ for convenience. Fig. \ref{fig_param_sensitivity} presents the average performance across five cross-domain scenarios for each dataset with varying parameter values. The results reveal that our model exhibits relative stability across a range of parameter values. Notably, the effect of $\gamma_1$ varies across datasets. On UCIHAR, performance remains consistent, while MFD benefits from larger values, and SSC achieves optimal performance around $0.2$. Similarly, $\gamma_2$ shows dataset-specific behavior. The optimal value is around $0.8$ for UCIHAR, and $0.6$ for MFD (with an initial increase followed by a decrease), and larger values tend to perform better on SSC, although sensitivity is lower overall. Interestingly, $\gamma_3$ and $\beta$ display similar trends on SSC and MFD, with generally better performance at lower values. Conversely, on UCIHAR, larger $\gamma_3$ and smaller $\beta$ values lead to improved performance. Overall, the analysis suggests that the proposed method exhibits low parameter sensitivity within the chosen range.

\subsubsection{Impact of Masking Level}
\begin{figure}[t]
\centering
\includegraphics[width=0.65\linewidth]{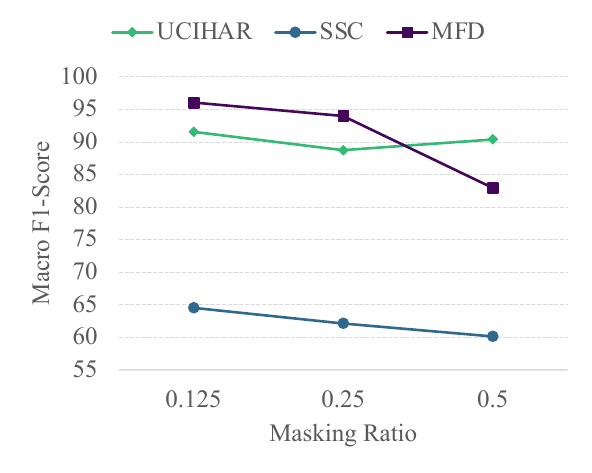}
\caption{Analysis of adaptation performance with different masking ratios.}
\label{fig_masking_ratio}
\end{figure}

Here, we systematically examine the impact of the masking ratio on adaptation performance within the context of imputation tasks. Specifically, we evaluate the performance of three different masking ratios ($12.5$\%, $25$\%, and $50$\%) on three benchmark datasets. The results presented in Fig. \ref{fig_masking_ratio} demonstrate improved performance with lower masking ratios in most cases. Notably, a masking ratio of $12.5$\% achieves the best performance on all datasets. These findings suggest that excessively masking the data during training can hinder adaptation performance in imputation tasks due to the increased difficulty imposed. 

\subsection{Calibration Evaluation Analysis}
\begin{figure*}[t]
\centering
\begin{tabular}{c}
\subfigure[]{\includegraphics[width=0.3\linewidth]{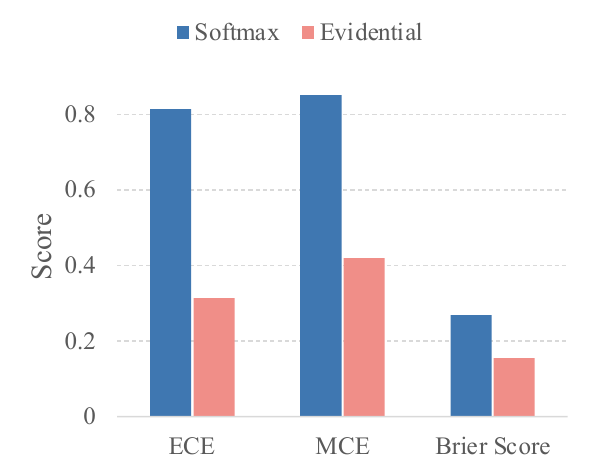} \label{fig_mapu_others_a}}
\subfigure[]{\includegraphics[width=0.3\linewidth]{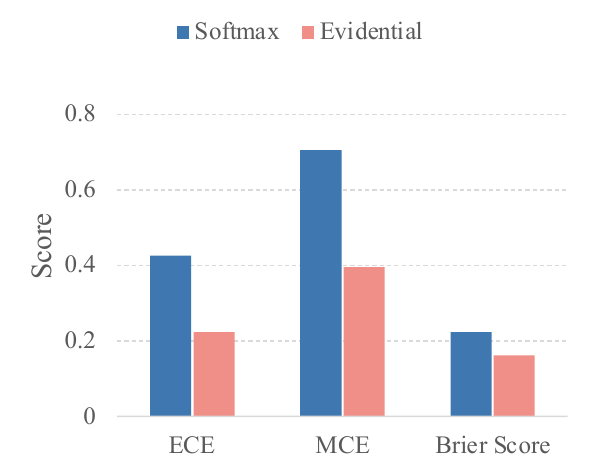} \label{fig_mapu_others_b}}
\subfigure[]{\includegraphics[width=0.3\linewidth]{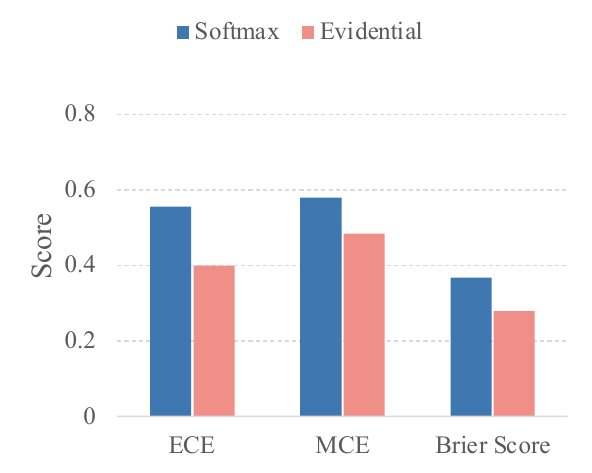} \label{fig_mapu_others_c}}

\end{tabular}
\caption{Average calibration errors of the pre-trained model with softmax and
evidential probabilities on three datasets. (a) UCIHAR dataset. (b) MFD dataset. (c) SSC dataset}
\label{fig_avg_calibration}
\end{figure*}

We evaluate the model's calibration performance using both softmax and evidential probabilities. Across three datasets and five cross-domain scenarios, we compute the average expected calibration error (ECE), maximum calibration error (MCE), and Brier scores (BS). Fig. \ref{fig_avg_calibration} summarizes the average calibration results for the UCIHAR, SSC, and MFD datasets. Detailed results for each cross-domain scenario are provided in the supplementary materials (i.e., Tables S1, S2, and S3).
The results consistently demonstrate superior calibration performance with the evidential uncertainty strategy. Compared to softmax predictions, evidential models achieve significantly lower ECE, MCE, and BS across all datasets. This suggests that softmax predictions are prone to overconfidence, leading to unreliable uncertainty estimates. In contrast, our proposed evidential learning strategy effectively mitigates miscalibration, resulting in a more reliable and better-calibrated model.

\subsection{Visualization Analysis}
\subsubsection{Robustness of the Evidential Uncertainty Metric}
\begin{figure}[h]
\centering
\begin{tabular}{c}
\subfigure[]{\includegraphics[width=0.49\linewidth]{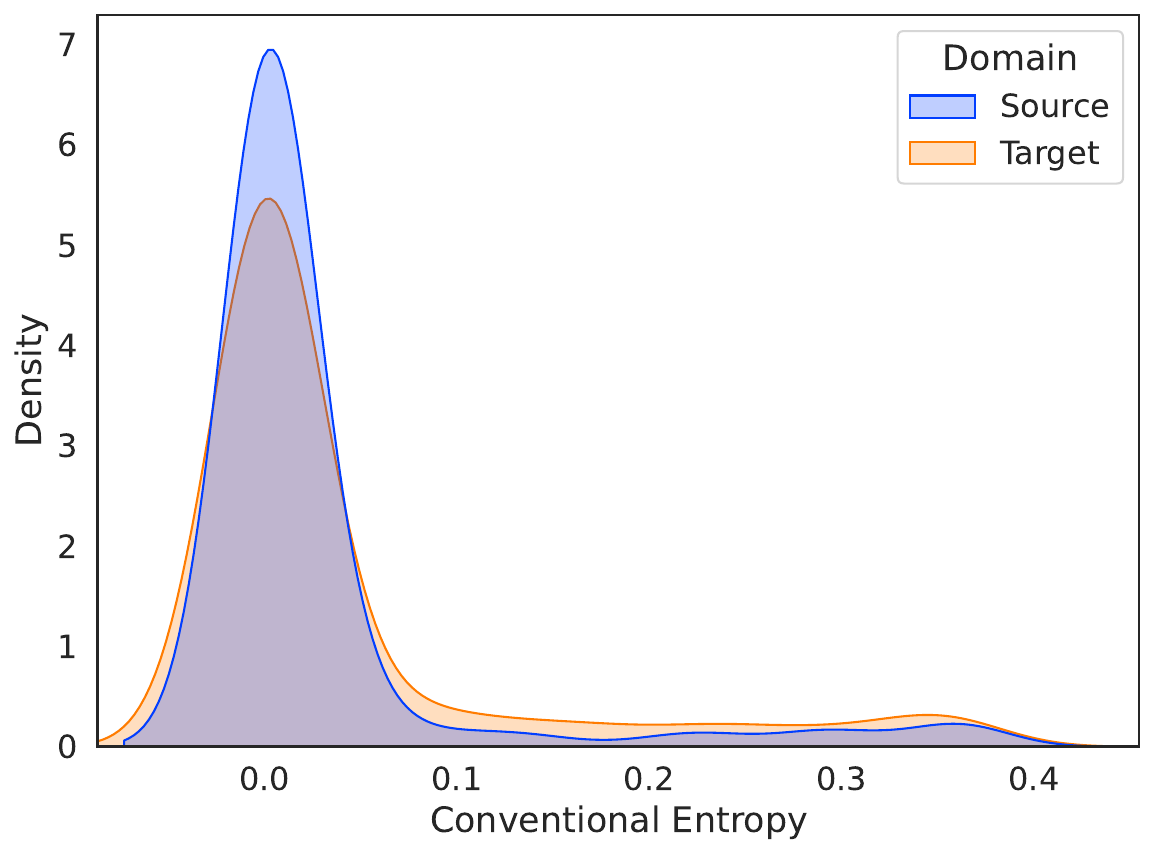} \label{fig_evidential_ood_a}}
\subfigure[]{\includegraphics[width=0.49\linewidth]{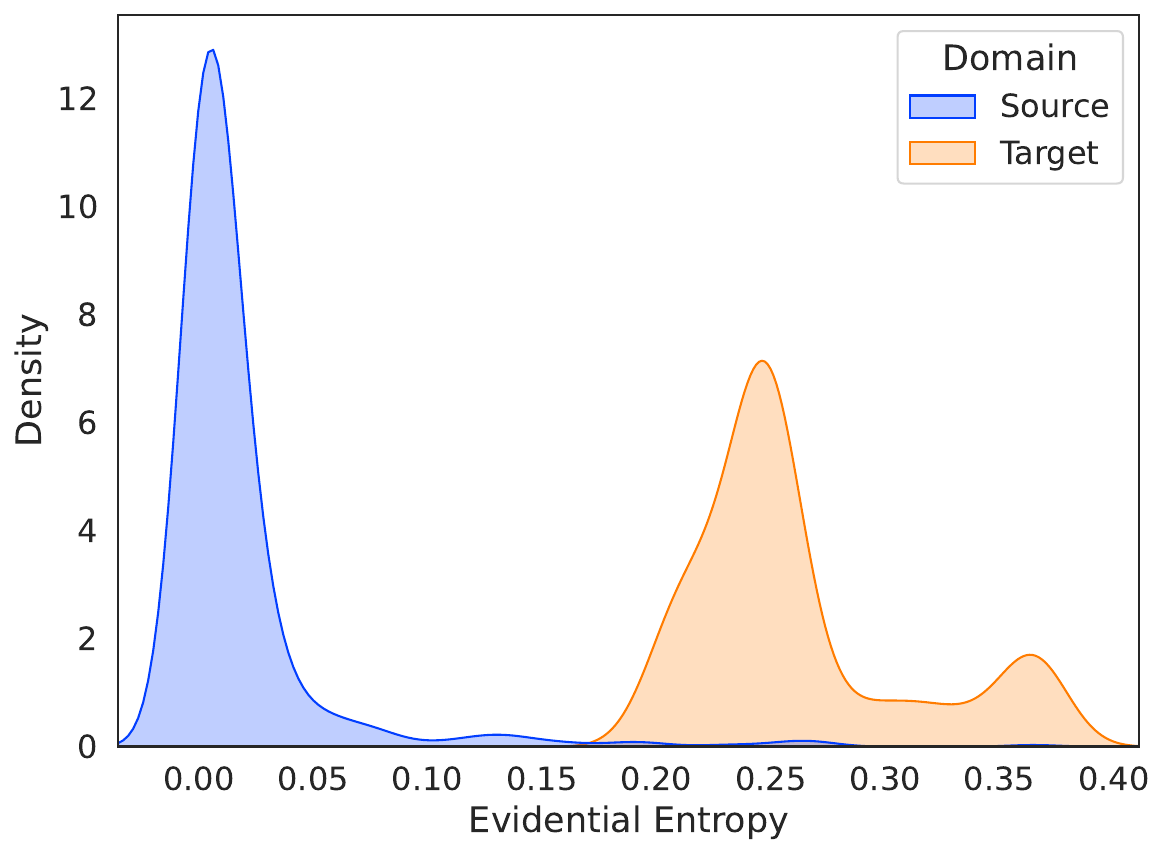} \label{fig_evidential_ood_d}} 
\\
\subfigure[]{\includegraphics[width=0.49\linewidth]{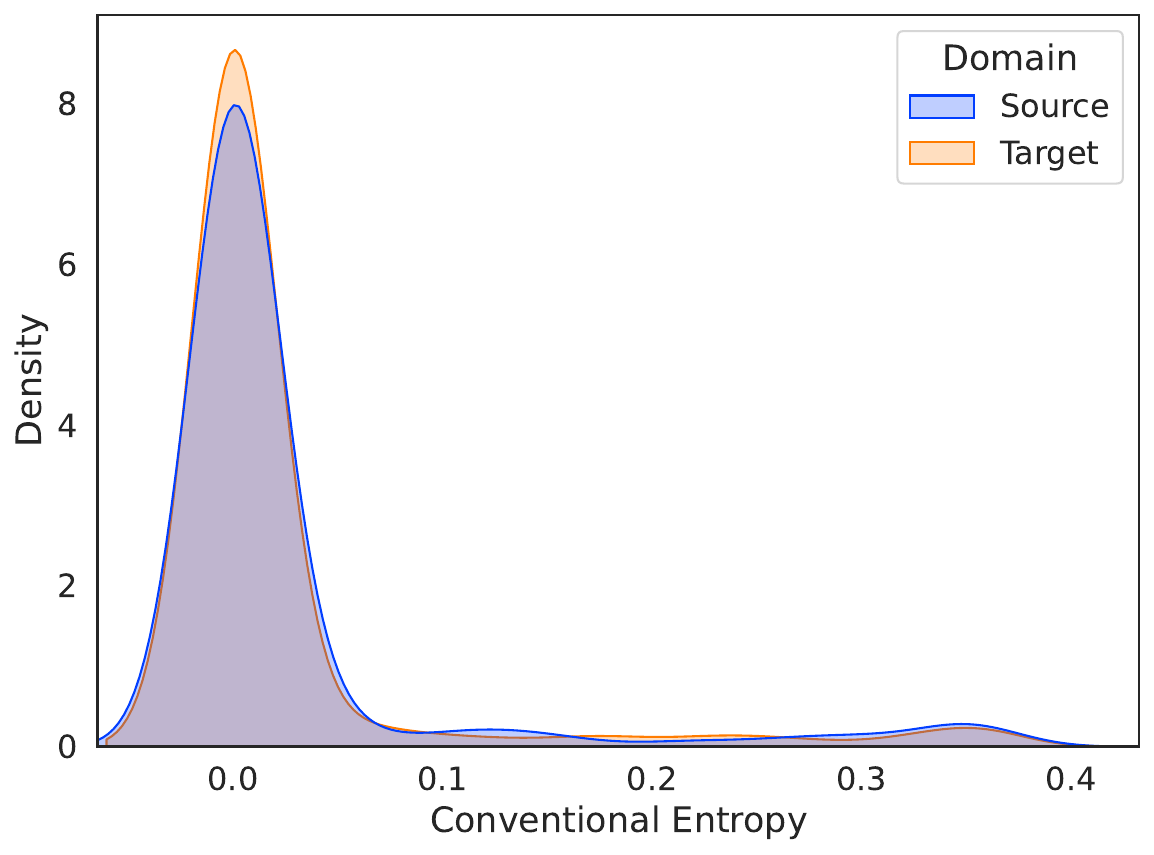} \label{fig_evidential_ood_b}}
\subfigure[]{\includegraphics[width=0.49\linewidth]{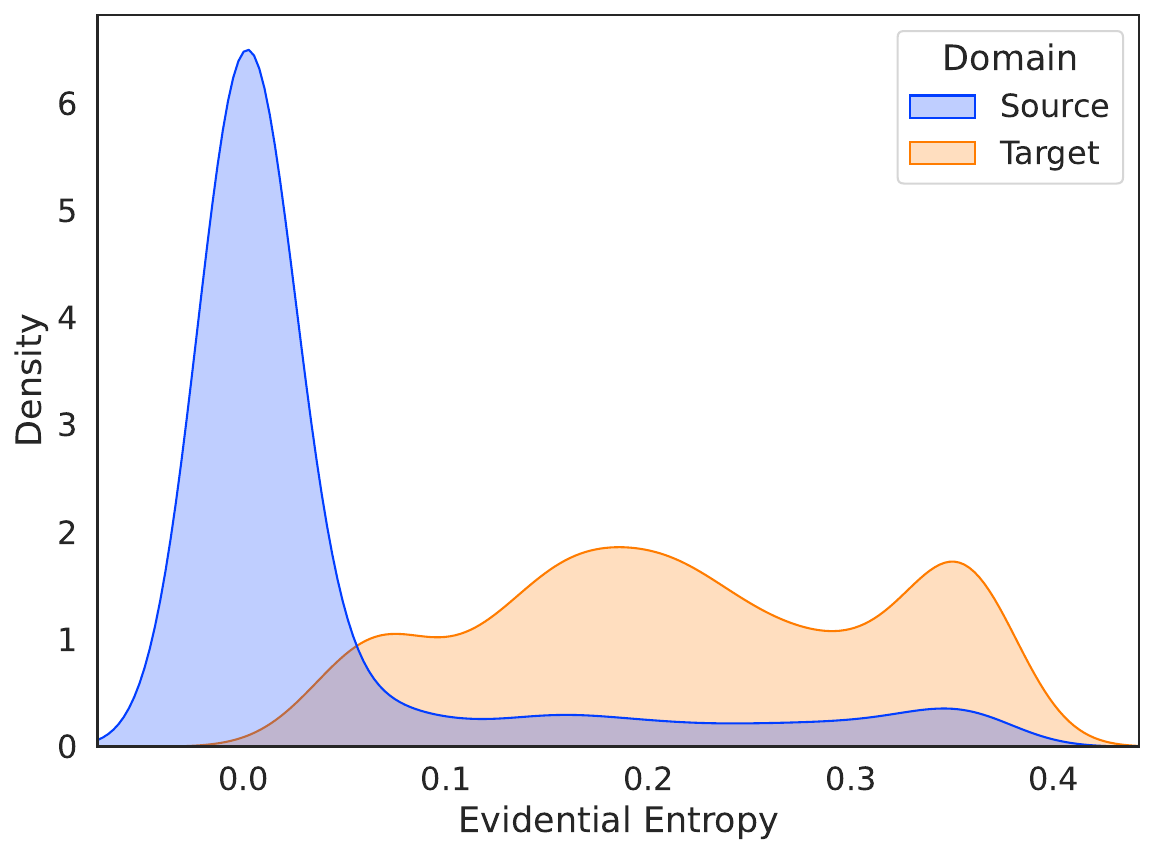} \label{fig_evidential_ood_e}}
\\
\subfigure[]{\includegraphics[width=0.49\linewidth]{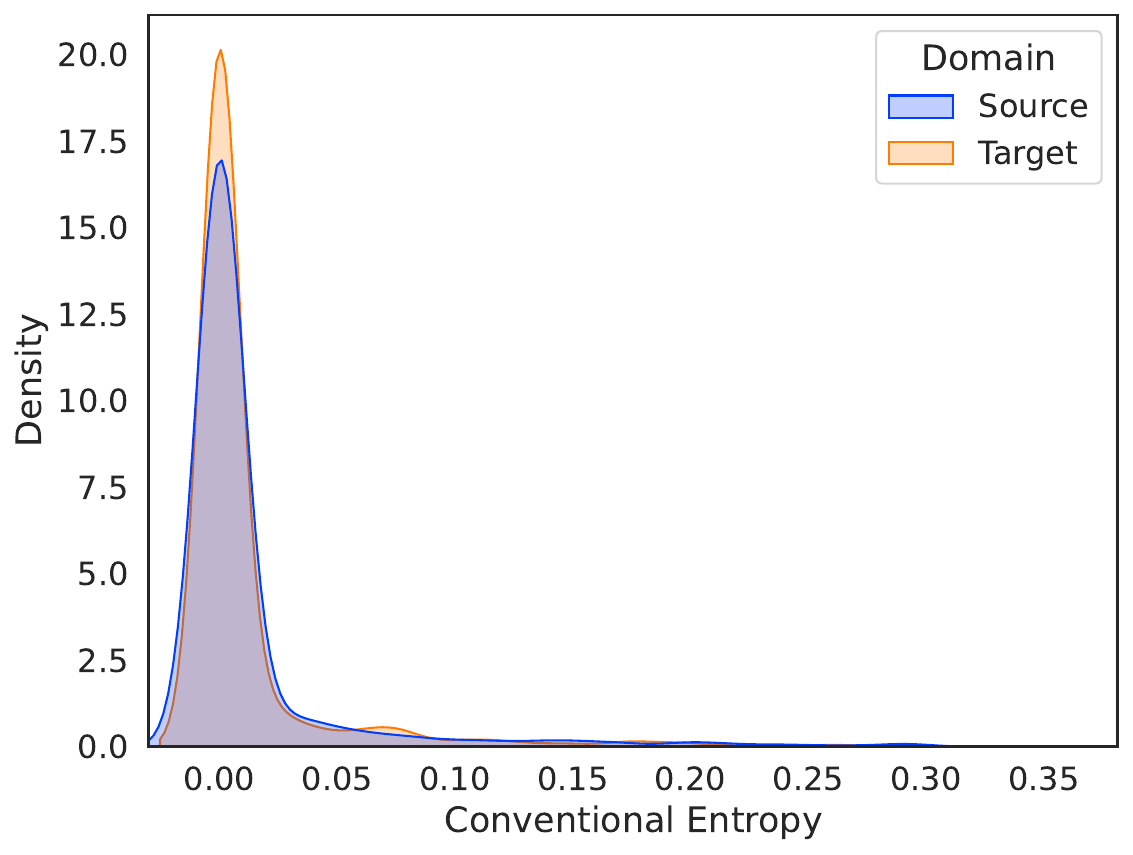} \label{fig_evidential_ood_c}}
\subfigure[]{\includegraphics[width=0.49\linewidth]{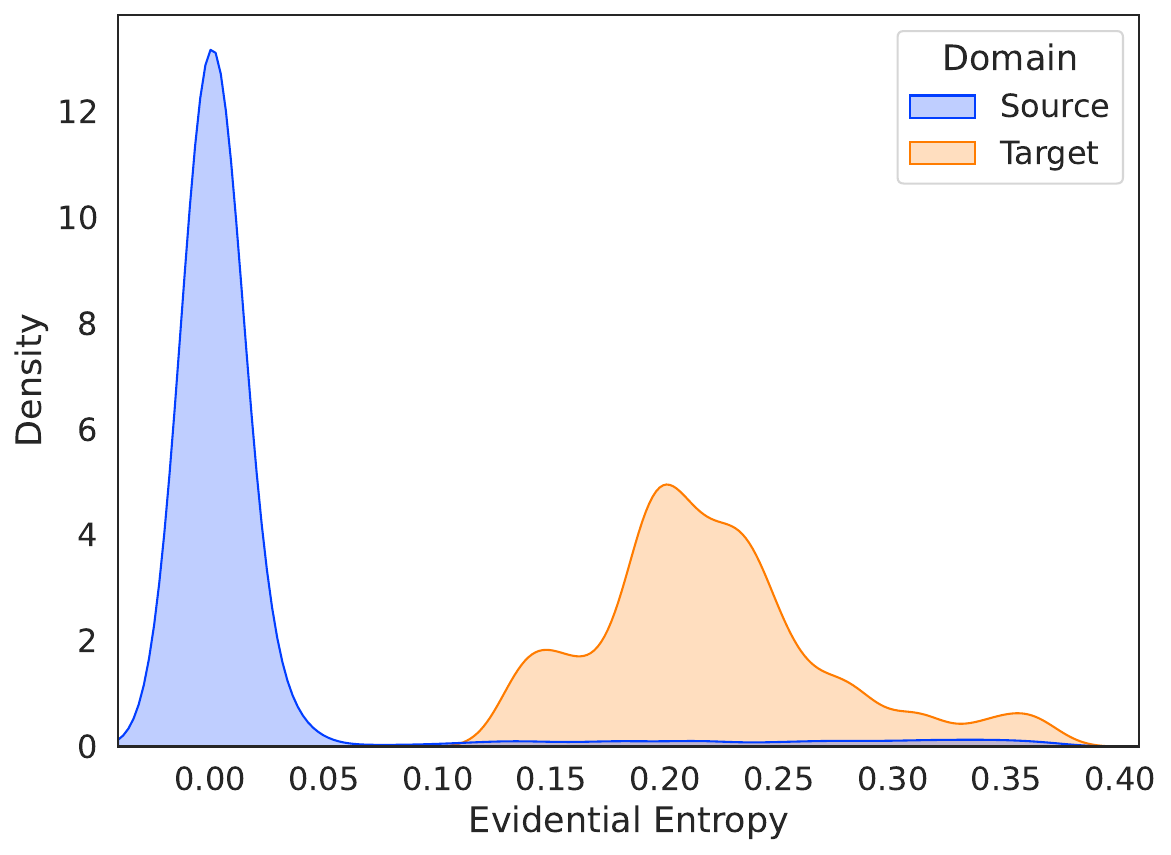} \label{fig_evidential_ood_f}}
\end{tabular}
\caption{Visualization of predictive entropy versus evidential entropy on different scenarios of three datasets (a) On the UCIHAR dataset. (b) On the SSC dataset. (c) On the MFD dataset.}
\label{fig_evidential_ood}
\end{figure}

This section evaluates the robustness of our proposed evidential uncertainty metric against the traditional softmax entropy. We measure the entropy of in-domain and out-of-domain data using both softmax probabilities and the proposed evidence-based probabilities on the UCIHAR, SSC, and MFD datasets. Fig. \ref{fig_evidential_ood} illustrates the estimated entropy values of one cross-domain scenario on each dataset for the source and target data. Visualizations of each cross-domain scenario on the three datasets are presented in the supplementary materials (i.e., Figures S1, S2, and S3). The results reveal a clear distinction between the two approaches. Conventional entropy exhibits consistently low values, even for out-of-domain data, indicating a problem of overconfidence. Conversely, our proposed evidential entropy demonstrates a clear pattern of differentiation between the source and target domains. This characteristic makes it more suitable for source-free adaptation tasks. Minimizing the evidential entropy of target samples during feature encoder optimization fosters better feature alignment between source and target domains. Consequently, this leads to improved model adaptation capabilities.

\subsubsection{Visualization of Feature Learning and Adaptation}
\begin{figure*}[t]
\centering
\begin{tabular}{c}
\subfigure[]{\includegraphics[width=0.3\linewidth, cfbox=gray 0.1pt -2pt]{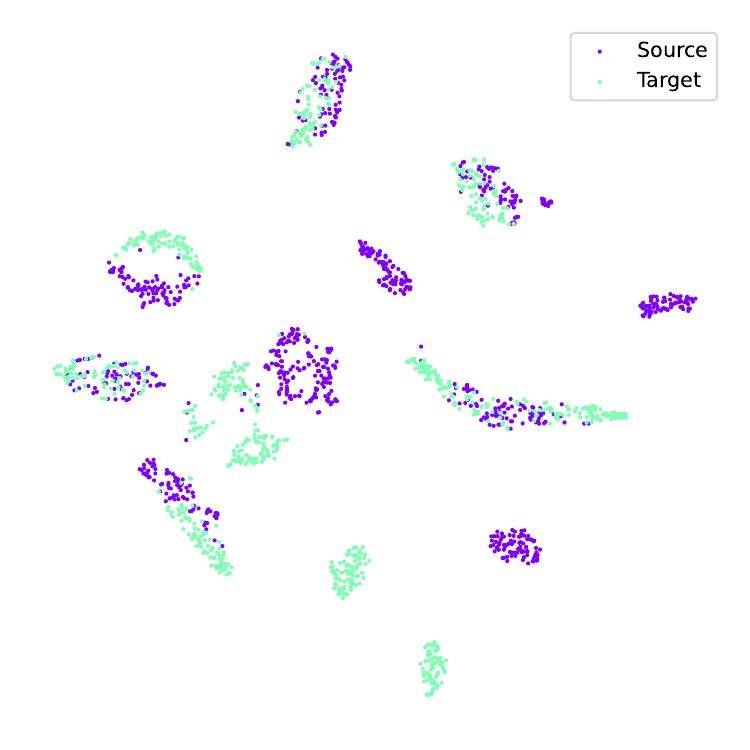} \label{fig_tsne_adaptation_a}}
\subfigure[]{\includegraphics[width=0.3\linewidth, cfbox=gray 0.1pt -2pt]{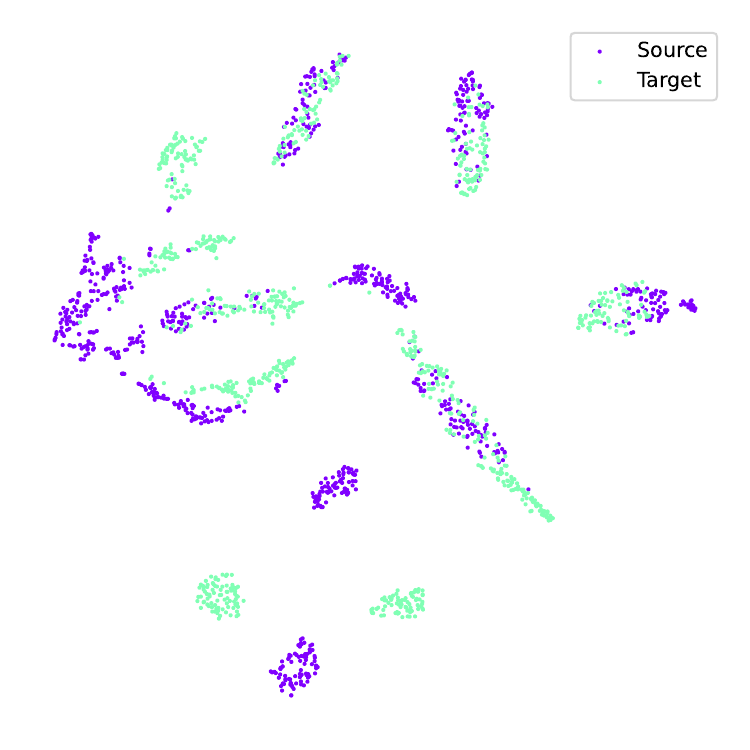} \label{fig_tsne_adaptation_b}}
\subfigure[]{\includegraphics[width=0.3\linewidth, cfbox=gray 0.1pt -2pt]{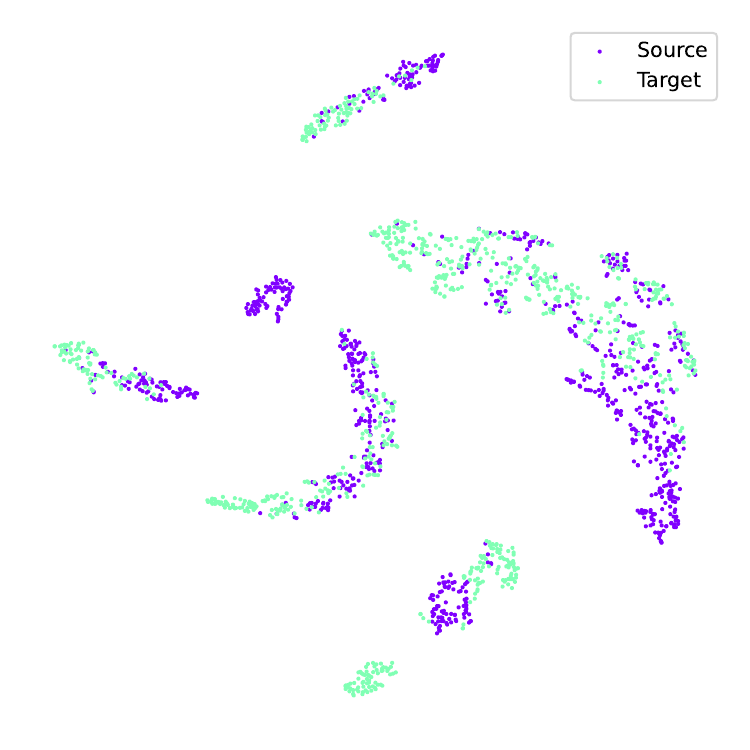} \label{fig_tsne_adaptation_c}}
\end{tabular}
\caption{The t-SNE feature visualizations of the source domain and target domain on the MFD dataset. (a) Source-model only. (b) MAPU. (c) E-MAPU}
\label{fig_tsne_adaptation}
\end{figure*}

We leverage t-SNE visualizations on the MFD dataset (Fig.~\ref{fig_tsne_adaptation}) to demonstrate the effectiveness of MAPU and E-MAPU in aligning target domain features with the source domain. Additional visualizations of UCIHAR and SSC are illustrated in the supplementary materials (i.e., Figures S4 and S5). First, MAPU achieves improved alignment compared to a source-only model, a benefit attributed to its temporal adaptation capabilities. Further, E-MAPU enhances this alignment due to its well-calibrated model, resulting from the evidential uncertainty optimization strategy. These observations suggest that during adaptation, the model utilizes evidential entropy to distinguish between source and target samples. Specifically, it minimizes the evidential entropy of out-of-support target samples (those outside the source domain's support). This is achieved by adapting the target encoder to map these samples to a new feature representation closer to the source domain's support, ultimately fostering better alignment and enhancing adaptation performance.

\section{Conclusion}
This paper introduces MAsk And imPUte (MAPU), a novel method for source-free domain adaptation (SFDA) on time series data. MAPU tackles the challenge of maintaining temporal consistency in time series data by proposing a novel temporal imputation task. This task focuses on recovering the original signal within the feature space, rather than the raw input space. Notably, MAPU is the first approach to explicitly account for temporal dependencies in a source-free SFDA setting for time series data. Furthermore, we enhance MAPU by incorporating evidential uncertainty learning, resulting in E-MAPU. This extension fosters a well-calibrated model and addresses the issue of overconfidence inherent in softmax predictions. The efficacy of both MAPU and E-MAPU is rigorously evaluated through extensive experiments on five real-world datasets. These experiments demonstrate significant performance gains compared to existing methods. Overall, this work underscores the potential of the proposed methods for mitigating domain shift problems in time series applications while preserving data privacy.

\bibliographystyle{IEEEtran}
\bibliography{main_arxiv}

\vfill

\end{document}